\documentclass[10pt,twocolumn,letterpaper]{article}

\usepackage[pagenumbers]{cvpr} 
\usepackage[accsupp]{axessibility} 
\usepackage[dvipsnames]{xcolor}
\newcommand{\red}[1]{{\color{red}#1}}

\definecolor{cvprblue}{rgb}{0.21,0.49,0.74}
\usepackage[pagebackref,breaklinks,colorlinks,citecolor=cvprblue]{hyperref}
\usepackage{multirow}
\usepackage{colortbl}  
\usepackage{xcolor}
\usepackage{svg}
\def\paperID{3537} 
\def\confName{CVPR}
\def\confYear{2024}

\definecolor{myblue}{HTML}{1f77b4}
\definecolor{myred}{HTML}{d62728}
\definecolor{boxred}{HTML}{AD2F2D}
\definecolor{mygreen}{HTML}{2ca02c}
\definecolor{myorange}{HTML}{ff7f0e}
\definecolor{mybrown}{HTML}{964d22}
\definecolor{nblgreen}{HTML}{009B5C}
\definecolor{Light}{RGB}{255,234,227}
\usepackage{utfsym}
\usepackage{wrapfig}
\usepackage{soul}
\usepackage{marvosym}
\title{Continual Forgetting for Pre-trained Vision Models}
\usepackage{mathrsfs}
\newlength\savewidth\newcommand\shline{\noalign{\global\savewidth\arrayrulewidth
  \global\arrayrulewidth 1pt}\hline\noalign{\global\arrayrulewidth\savewidth}}
  
\author{
    Hongbo~Zhao$^{1,3}$\thanks{Equal contribution.} \hspace{0.8mm}
    Bolin~Ni$^{1,3*}$\hspace{0.8mm}
    Junsong~Fan$^{2}$ \hspace{0.8mm}
    Haochen~Wang$^{1,3}$ \hspace{0.8mm}
    Yuxi~Wang$^{2}$ \hspace{0.8mm}\\
    Fei~Zhu$^{2}$ \hspace{0.8mm}
    Yuntao~Chen$^{2}$ \hspace{0.8mm}
    Gaofeng~Meng$^{1,2,3}\thanks{Corresponding author.}$\hspace{0.8mm}
    Zhaoxiang~Zhang$^{{1,2,3,4}\dag}$\\[8pt]
    \small{$^1$
State Key Laboratory of Multimodal Artificial Intelligence Systems, Institute of Automation, Chinese Academy of Sciences} \\
\small{$^2$Centre for Artificial Intelligence and Robotics, Hong Kong Institute of Science \& Innovation, Chinese Academy of Sciences} \hspace{3mm} \\
   \small{$^3$ 
    University of Chinese Academy of Sciences}
    \hspace{3mm}
    \small{$^4$ Shanghai Artificial Intelligence Laboratory} \\
    \small{\texttt{\{zhaohongbo2022, zhaoxiang.zhang\}@ia.ac.cn}} \quad
    \small{\texttt{gfmeng@nlpr.ia.ac.cn}}
}

\begin{document}
\maketitle
\begin{abstract}
For privacy and security concerns, the need to erase unwanted information from pre-trained vision models is becoming evident nowadays.
In real-world scenarios, erasure requests originate \emph{at any time} from both users and model owners.
These requests usually form a sequence.
Therefore, under such a setting, selective information is expected to be \emph{continuously} removed from a pre-trained model while maintaining the rest.
We define this problem as continual forgetting and identify two key challenges.
\textbf{\textit{(i)}} For unwanted knowledge, efficient and effective deleting is crucial.
\textbf{\textit{(ii)}} For remaining knowledge, the impact brought by the forgetting procedure should be minimal. 
To address them, we propose \textbf{G}roup \textbf{S}parse \textbf{LoRA} (GS-LoRA).
Specifically, towards \textbf{\textit{(i)}},  we use LoRA modules to fine-tune the FFN layers in Transformer blocks for each forgetting task independently, and
towards \textbf{\textit{(ii)}}, 
a simple group sparse regularization is adopted, enabling automatic selection of specific LoRA groups and zeroing out the others.
GS-LoRA is effective, parameter-efficient, data-efficient, and easy to implement.
We conduct extensive experiments on face recognition, object detection and image classification and demonstrate that GS-LoRA manages to forget specific classes with minimal impact on other classes.
Codes will be released on \url{https://github.com/bjzhb666/GS-LoRA}.
\end{abstract}    
\section{Introduction}
\label{sec:intro}


As 
pre-trained models become larger nowadays, more training data are required.
These data are usually collected through various ways such as the Internet, books, publicly available datasets, and manual labeling.
Within the vast amount of data, there is often erroneous or privacy-sensitive information and pre-trained models may learn from it.
For instance, the ImageNet Roulette project~\cite{crawford2021excavating, ImageNet78:online} shows models tend to be biased toward racist, misogynistic, and cruel \etc.
%
Furthermore, with increased public awareness of privacy protection and updated privacy regulations~\cite{regulation2018general,goldman2020introduction}, individuals are now demanding the removal of any privacy-related information immediately.
Therefore, practical model erasing techniques are required upon receiving a deletion request.
%
\begin{figure}[t]
  \centering
   \includegraphics[width=\linewidth]{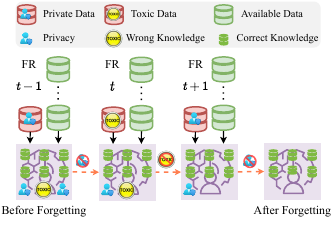}
   \caption{
   \textbf{Illustration of continual forgetting},
   which aims to remove specific knowledge in pre-trained models sequentially.
   ``FR'' stands for Forgetting Request.
   The \red{red} data (privacy data, toxic data, \etc) contains unwanted knowledge which needs to be removed, while the \textcolor{mygreen}{rest} should be maintained.
    The model {inherits parameters} from the last forgetting task at the beginning of a new forgetting task.
}
   \label{fig: motivation}
   \vspace{-10pt}
\end{figure}
In real-world scenarios,  these requests usually originate at any time from both users and model owners and naturally form a sequence. 
Under such a setting, selective information is expected to be \textit{continuously} removed from a pre-trained model while maintaining the rest. 
We identify this novel problem as \textbf{continual forgetting} and illustrate it in \cref{fig: motivation}, where privacy and wrong knowledge need to be removed from the pre-trained model sequentially.
This task holds significance in upholding privacy and reducing unwanted model bias, such as gender and racial discrimination, in real-world applications.

 
A related research topic is machine unlearning, which refers to the process of removing or erasing knowledge or patterns that a machine learning model has learned during training.
Prior attempts mostly focused on typical machine learning algorithms~\cite{mahadevan2021certifiable,bourtoule2021machine,chen2019novel,izzo2021approximate}, \eg, linear/logistic regression~\cite{mahadevan2021certifiable,izzo2021approximate}, and thus have a limited scope of application.
Recent studies that explored unlearning techniques for deep learning models are either computationally heavy and only effective on small-scale problems~\cite{guo2019certified,golatkar2020eternal,sekhari2021remember}, or require specific designs in the pre-training process~\cite{bourtoule2021machine, yan2022arcane}, which are impractical. 
These approaches lack the ability to proceed with numerous everyday requests and thus are not capable of continual forgetting.
We identify two key challenges in the design of continual forgetting algorithms.
\textbf{\textit{(i)}} Efficient and effective deleting for unwanted knowledge is crucial.
Especially for continual forgetting scenarios, lightweight and fast modifications are more important to achieve deleting information promptly.
\textbf{\textit{(ii)}} Should have minimal impact on remaining knowledge, \ie, catastrophic forgetting should be mitigated.

To this end, we propose \textbf{G}roup \textbf{S}parse \textbf{LoRA} (GS-LoRA).
Specifically, to achieve efficient forgetting on unwanted knowledge, we utilize LoRA~\cite{hu2021lora} to fine-tune the FFN modules in Transformer blocks inspired by parameter-efficient fine-tuning (PEFT) techniques~\cite{hu2021lora,houlsby2019parameter,li2021prefix} and Geva \etal~\cite{geva2020transformer}.
To mitigate catastrophic forgetting on remaining knowledge~\cite{kirkpatrick2017overcoming}, 
we use a group sparse regularizer to achieve a sparse and accurate modification of FFN modules, as fine-tuning fewer parameters is observed to be effective~\cite{mallya2018piggyback, Mallya_Lazebnik_2018,l2p,zhang2022continual} towards alleviating catastrophic forgetting.
This is akin to conducting minimally invasive surgery on a model instead of a major surgery.
%
GS-LoRA is effective, parameter-efficient, data-efficient, easy to implement, and applicable to large models.
To verify the effectiveness of our proposed GS-LoRA, we initially conduct experiments on face recognition because it is a fundamental privacy-sensitive task, and then evaluate it on a more general task, \ie, object detection.
Empirically, GS-LoRA performs well in both settings, indicating that our method is a general framework with minimal domain knowledge and few inductive biases across various vision tasks.

Our contributions are summarized as follows:
\begin{itemize}
\item We are the first to propose the continual forgetting problem, which is essential in practical scenarios for fast model editing and privacy protection.

\item 
To address this problem, we first identify two challenges and propose GS-LoRA to achieve efficient and effective forgetting while maintaining the performance of the rest.
\item Extensive experiments on both face recognition and object detection demonstrate that GS-LoRA effectively forgets specific classes while maintaining high performance on the remaining categories.
\end{itemize}


\section{Related Work}
\label{sec:related work}


\subsection{Continual Learning}

Continual learning aims to enable models to acquire new knowledge without forgetting previously learned information \cite{kirkpatrick2017overcoming}.
It is a learning paradigm that is particularly applicable in dynamic and changing scenarios.
Researchers have designed three strategies to achieve this goal, including rehearsal-based methods \cite{lavda2018continual,rebuffi2017icarl,isele2018selective,liu2023continual,rolnick2019experience,shin2017continual,zhu2021prototype,zhu2022learning,chen2023diffusepast}, regularization-based methods \cite{kirkpatrick2017overcoming,li2017learning,aljundi2018memory,schwarz2018progress,zhu2021class}, and structure-based methods \cite{mallya2018piggyback,zhang2022continual,rusu2016progressive,Douillard_Rame_Couairon_Cord_2022,aljundi2017expert,liu2021adaptive}.
These three strategies for continual learning are frequently combined to improve performance \cite{rebuffi2017icarl,liu2023continual,zhu2023imitating,Ni_2024_CVPR}.

Our proposed GS-LoRA falls into the category of structure-based methods.
However, our problem differs from continual learning as we aim to continuously delete, rather than add new knowledge to the model. 

\subsection{Machine Unlearning}
\label{MUL}


Machine unlearning involves retraining or modifying machine learning models to diminish or eradicate the influence of previously acquired patterns or biases, aiming to enhance the models' fairness and safety \cite{nguyen2022survey, xu2023machine, bourtoule2021machine, shibata2021learning, brophy2021machine,ginart2019making}.
A lot of studies design unlearning algorithms on simple machine learning algorithms \cite{mahadevan2021certifiable,izzo2021approximate,baumhauer2022machine,chen2019novel,sun2023lazy,brophy2021machine}. As a result, the applicability of these algorithms is constrained.
Initial work on forgetting in deep learning either slices the data and trains a series of submodels to isolate the effect of specific data points on the model \cite{bourtoule2021machine, yan2022arcane,shibata2021learning} (exact unlearning) or calculates influence functions to approximate the impact of a data item on the parameters of models \cite{golatkar2020eternal,guo2019certified,sekhari2021remember, jang2022knowledge} (approximate unlearning).
However, these methods deteriorate when applied to larger datasets and models, and the computational cost is exceedingly high.




Our problem focuses on the continual forgetting of a pre-trained model. 
One previous work \cite{shibata2021learning} studies the continual exact unlearning by adding a class-specific synthetic signal in the pre-training stage.
It should be noted that specific designs cannot be performed in the pre-training process, which is not common in deep learning applications.
Cha~\etal \cite{cha2023learning} mentions instance-wise forgetting and its continual form, while our setting is at category-level.

\subsection{Parameter-Efficient Fine-Tuning}

Training large models by self-supervised learning and then fine-tuning them on downstream tasks has become a new paradigm of deep learning \cite{gpt3,gpt2, He2022,he2020momentum,radford2021clip,li2023blip2, wang2023hard, wang2023droppos, wang2023bootstrap,ni2022expanding}.
Parameter-efficient fine-tuning (PEFT) techniques \cite{hu2021lora,li2021prefix,houlsby2019parameter,jia2022visual,samadapter,zhou2022conditional, Tip-adapter} are proposed to optimize a limited number of parameters, as fully fine-tuning increasing large models \cite{gpt2,gpt3,kirillov2023segment,MIR} becomes less practical for various downstream tasks.

Recent studies focus on three different types of PEFT methods, categorized based on the origin of trainable parameters. 
These methods include addition-based approaches \cite{li2021prefix, houlsby2019parameter, liu2021gpt}, 
freezing-based techniques \cite{ravfogel2021bitfit,lee2019would}, 
and parameter-factorization-based methods \cite{hu2021lora,valipour2022dylora, chavan2023one}. 
All these methods are designed to improve the performance of downstream tasks, while our method modifies pre-trained models with the help of PEFT.

\section{Problem Setting}
We propose a new problem termed continual forgetting, which involves the selective removal of specific knowledge from a pre-trained model while preserving the performance of the rest.
In this section, we first consider the simplest situation where there is only one task that needs to be forgotten, and later extend to a continual form.

Let $\mathcal{M}$ be a model pre-trained on the dataset $D$, we denote the mapping relationship of the model as $f_M:\mathcal{X}_D \rightarrow \mathcal{Y}_D$, where $\mathcal{X}_D$ and $\mathcal{Y}_D$ represent the input set and output set, respectively. 
Our objective is to selectively discard certain knowledge in the model while retaining the rest.
Let $D_f$ and $D_r$ represent datasets containing knowledge to be forgotten and retained.
Given that $|D_r|$ is typically large in practical scenarios and the retraining process is time-consuming, we require $|D_r|+|D_f|\ll |D|$.
Before forgetting, model $\mathcal{M}$ performs well on both $D_f$ and $D_r$, \ie, 
\begin{equation}
   f_M:
   \mathcal{X}_{D_f} \stackrel{f_M}{\longrightarrow} \mathcal{Y}_{D_f}, \mathcal{X}_{D_r} \stackrel{f_M}{\longrightarrow} \mathcal{Y}_{D_r}. 
\end{equation}
The forgetting algorithm $\mathscr{F}$ modifies the model to obtain $\mathcal{M}' = \mathscr{F}(\mathcal{M},D_f,D_r)$ and a new mapping relationship $f_{M'}$ satisfying
\begin{equation}
f_{M'}:   \mathcal{X}_{D_f} \not \stackrel{f_{M'}}{\longrightarrow} \mathcal{Y}_{D_f}, \mathcal{X}_{D_r}  \stackrel{f_{M'}}{\longrightarrow} \mathcal{Y}_{D_r}.
\end{equation}
Here, $\not \stackrel{f_{M'}}{\longrightarrow}$ means the mapping relationship no longer holds.

Now, we extend the problem to a continual form where the model is required to sequentially forget specific knowledge.
Let $D_r=\{D_{r_t}\}$ and $D_f=\{D_{f_t}\}$ for $t=1,2,\cdots, T$ represent two sequences of datasets, where $T$ is the number of forgetting tasks, $D_{{f_t}/{r_t}}=\{(x_{{f_t}/{r_t}}^i,y_{{f_t}/{r_t}}^i)_{i=1}^{n_t} \}$ is the forgotten or retained dataset of the $t$-th task, $x_{{f_t}/{r_t}}^i \in \mathcal{X}_{{f_t}/{r_t}}$ is an input and $y_{{f_t}/{r_t}}^i \in \mathcal{Y}_{{f_t}/{r_t}}$ is the corresponding label.
The forgetting algorithm $\mathscr{F}$ handles erase requests sequentially, starting from $\mathcal{M}$, and generates a sequence of models $\mathcal{M}_{f_1},\mathcal{M}_{f_2},\cdots,\mathcal{M}_{f_t},\cdots,\mathcal{M}_{f_T}$, where $\mathcal{M}_{f_t}$ represents the modified model after the $t$-th forgetting task.
After processing task $\mathcal{T}_t$, model $\mathcal{M}_{f_t}$ performs poorly on ${D_{f_i}}$ but maintains the original performance in the remaining part, \ie, the corresponding mapping relationship $f_{M_t}$ holds 
\begin{equation}\label{eq:clforget}
f_{M_t}:
    \mathcal{X}_{D_{f_i}} \not \stackrel{f_{M_t}}{\longrightarrow} \mathcal{Y}_{D_{f_i}}, \mathcal{X}_{D_{r_t}}  \stackrel{f_{M_t}}{\longrightarrow} \mathcal{Y}_{D_{r_t}},
\end{equation} 
where $i=1,2,\cdots,t, t=1,2,\cdots,T.$

\section{Method}

\begin{figure}
    \centering
    \includegraphics[width=\linewidth]{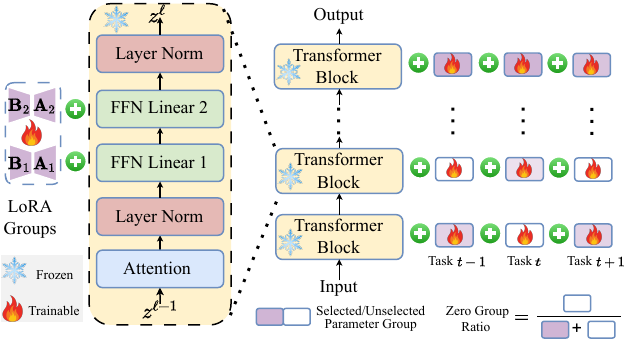}
    \caption{\textbf{Overall pipeline of GS-LoRA.}
    We incorporate a set of LoRA modules in each continual forgetting task and adopt a sparse structure selection strategy to achieve accurate and few modifications.
    All LoRA modules are added in the Linear layers of FFN in the Transformer blocks and we regard the LoRA modules in a Transformer block as one group.
    We use group sparse regularization to automatically select LoRA groups.
    The purple groups are selected to modify and the white groups are neglected.
    The pre-trained model (including Transformer blocks and other parts) is frozen and only LoRA groups are trainable.}
    \label{fig:3}
    \vspace{-0pt}
\end{figure}

\textbf{Preliminary: LoRA.}    Hu \etal \cite{hu2021lora} argue that the weight matrix in the pre-trained model has a very low intrinsic rank and utilizes a low-rank decomposition to implement parameter updates.
For a weight matrix $\mathbf{W}\in \mathbb{R}^{d\times k}$, it is updated following $\mathbf{W} = \mathbf{W} + \Delta \mathbf{W} = \mathbf{W} + \mathbf{B}\mathbf{A}$, where $\mathbf{B}\in \mathbb{R}^{d \times r}$ and $\mathbf{A} \in \mathbb{R}^{r \times k}$ are low rank matrices and $r \ll min\{d,k\}$ is the rank of matrix $\mathbf{B}$ and $\mathbf{A}$.
Only matrices with low ranks are trainable, while the matrix $\mathbf{W}$ remains frozen during training.
LoRA can be added to the linear projection matrices in Multi-Head Attention modules or the Feed-Forward Network (FFN) modules in Transformer blocks.

\subsection{Overview}
Considering two key challenges in \cref{sec:intro} and the optimization goal in \cref{eq:clforget}, we propose Group Sparse LoRA (GS-LoRA) with selective forgetting loss and knowledge retention loss to achieve continual forgetting.
\cref{fig:3} shows the overall pipeline of GS-LoRA.
To achieve efficient forgetting, we use LoRA to fine-tune the FFN modules in Transformer blocks. 
To mitigate catastrophic forgetting of the remaining knowledge, smaller network changes are preferred  \cite{Mallya_Lazebnik_2018,l2p,zhang2022continual,mallya2018piggyback}.
Therefore, we use group sparse regularization to select and modify fewer blocks. 
\cref{GS-LoRA} gives a more detailed description.
To achieve the optimization goal in \cref{eq:clforget}, we use a selective forgetting loss to maximize the original loss for the forgotten classes and a knowledge retention loss to minimize the loss for the remaining classes in \cref{lossdata}.

\subsection{GS-LoRA}\label{GS-LoRA}

\textbf{LoRA Based Model Tuning.}
Following the findings of Geva \etal \cite{geva2020transformer}, FFN layers in the Transformer blocks store a substantial amount of knowledge, necessitating modification of the FFN modules to achieve knowledge erasure.
Although directly modifying these layers is theoretically feasible, it is inefficient due to the large number of parameters in the FFN layers.
To reduce the learnable parameters, we incorporate a set of LoRA modules to the FFN in each Transformer block and only make these LoRA modules trainable.

Suppose $\mathbf{x}$ is the input of the $\ell$-th FFN module, the mathematical form can be expressed as: 
\begin{equation}
    \mathcal{FFN}^{(\ell)}(\mathbf{x}) = \max \left(\mathbf{0}, \mathbf{x}\mathbf{W}_1^{(\ell)} + \mathbf{b}_1^{(\ell)}\right)\mathbf{W}_2^{(\ell)} + \mathbf{b}_2^{(\ell)},
\end{equation} 
where $\mathbf{W}_1^{(\ell)}, \mathbf{W}_2^{(\ell)}, \mathbf{b}_1^{(\ell)}, \mathbf{b}_2^{(\ell)}$ are the weights and biases of two fully connected layers from the pre-trained model, respectively. 
We use LoRA to only fine-tune the weights of FFN modules:
\begin{equation}  
\begin{aligned}
\mathbf{W}^{(\ell)}_t&=
\begin{bmatrix}
    {\mathbf{W}_1}^{(\ell)}_{t} \\
    {\mathbf{W}_2}^{(\ell)}_{t}
\end{bmatrix}=
\begin{bmatrix}
    \mathbf{W}_1^{(\ell)} \\
    \mathbf{W}_2^{(\ell)}
\end{bmatrix}+
\sum_{i=1}^{t} \mathbf{B}^{(\ell)}_i \mathbf{A}^{(\ell)}_i,\\
\mathbf{B}^{(\ell)}_i&=
\begin{bmatrix}
    {\mathbf{B}_{1}}^{(\ell)}_i & \mathbf{O} \\
    \mathbf{O} &   {\mathbf{B}_{2}}^{(\ell)}_i
\end{bmatrix},\quad
\mathbf{A}^{(\ell)}_i=
\begin{bmatrix}
     {\mathbf{A}_{1}}^{(\ell)}_i  \\
       {\mathbf{A}_{2}}^{(\ell)}_i
\end{bmatrix},
\end{aligned}
\end{equation}
where ${\mathbf{W}_1}^{(\ell)}_{t}$ and ${\mathbf{W}_2}^{(\ell)}_{t}$ denote the  weights of the $\ell$-th FFN modules after task $\mathcal{T}_t$, and  ${\mathbf{B}_1}^{(\ell)}_i,{\mathbf{A}_1}^{(\ell)}_i,{\mathbf{B}_2}^{(\ell)}_i,{\mathbf{A}_2}^{(\ell)}_i$ for $i=1,2,\cdots,t$ refer to the corresponding LoRA matrices in task $\mathcal{T}_i$.
$\mathbf{O}$ is the zero matrix.
Note that the output FFN layers are frozen to ensure forgetting occurs in the backbone and is difficult to recover. 
A detailed discussion can be found in \cref{sec:6.1}.

\noindent\textbf{Group Sparsity Selection.}
To mitigate catastrophic forgetting and achieve precise modifications automatically, we introduce a group sparsity selection strategy that enables the selection of fewer Transformer blocks.
Although there are many ways to conduct a selection like routers \cite{yuksel2012twenty,shazeer2017outrageously}, meta learning \cite{vilalta2002perspective}, neural architecture search \cite{ren2021comprehensive,chen2019detnas}, 
we utilize \textit{group Lasso}, known for its simplicity and effectiveness in selecting parameters for specific groups \cite{yuan2006model, wen2016learning, feng2015learning,liu2015sparse} while setting others to zero.
Suppose LoRA matrices added to the $\ell$-th Transformer block in task $\mathcal{T}_t$ are ${\mathbf{B}_{1}}^{(\ell)}_t,{\mathbf{A}_{1}}^{(\ell)}_t,{\mathbf{B}_{2}}^{(\ell)}_t,{\mathbf{A}_{2}}^{(\ell)}_t$.
Then the optimization goal with group sparse regularization can be expressed as follows:
\begin{equation}\label{loss}
    \mathcal{L}_{total} = \mathcal{L}_{data}+\alpha \mathcal{L}_{strcuture}.
\end{equation}
Here, $\mathcal{L}_{data}$ denotes the loss on data, which will be elaborated in \cref{lossdata}, $\mathcal{L}_{structure}$ is the group sparse loss, and $\alpha$ serves as a hyperparameter to regulate the sparse intensity.

The group sparse loss on a set of weights can be represented as: 
\begin{equation}
    \mathcal{L}_{structure} =\sum_{\ell=1}^G \mathcal{L}_{gs}^{(\ell)},
\end{equation} 
where $G$ is the number of groups, $\mathcal{L}_{gs}^{(l)}$ is the group sparse loss of the $\ell$-th group.
We regard the LoRA weights in one Transformer block as a group. Therefore, the group sparse loss in the $\ell$-th group can be written as:
\begin{equation}
\mathcal{L}_{gs}^{(l)}=\Vert{\mathbf{B}}^{(\ell)}_t\Vert_F+\Vert{\mathbf{A}}^{(\ell)}_t\Vert_F.
\end{equation}
Here, $\Vert\cdot\Vert_F$ is the Frobenius norm of the LoRA matrices and $t$ denotes task $\mathcal{T}_t$.

\noindent
\textbf{Sparsity Warmup.}
Deep learning models tend to converge to local minima in the landscape \cite{poggio2017theory}.
When a high sparsity constraint is imposed, the model's ability to escape local minima is hindered, thus preventing the realization of forgetting.
However, achieving a sparse update necessitates a relatively large $\alpha$.
We adopt a warm-up strategy \cite{he2016deep} to address this conflict. 
We utilize a stepwise $\alpha$ to achieve effective forgetting while ensuring a sparse modification. 
The mathematical expression can be written as:
\begin{equation}\label{eq:warmup}
    \alpha_k=\left\{
\begin{aligned}
&0, &k < K,\\
&\alpha_{K},\quad &k\ge K.\\
\end{aligned}
\right.
\end{equation}
Here, $K$ is a hyperparameter. 
The model escapes the local minima in $k$ epochs without structure loss and then performs group sparsification to obtain a sparse modification.

\subsection{Loss Function} \label{lossdata}
In this section, we will discuss the data loss in \cref{loss} and introduce selective forgetting loss and knowledge retention loss to handle our continual forgetting problem.

\begin{table*}[t!]
\centering\small
\setlength{\tabcolsep}{5.5pt}
\begin{tabular}{ccccccccccccccc}
\toprule
\multirow{2}{*}{Methods} &\multirow{2}{*}{\begin{tabular}[c]{@{}c@{}}Tunable\\ Ratio $\downarrow$ \end{tabular}} & \multicolumn{3}{c}{100-5} & \multicolumn{3}{c}{100-10} & \multicolumn{3}{c}{100-50} & \multicolumn{3}{c}{100-90} \\ 
\cmidrule(lr){3-5} \cmidrule(lr){6-8} \cmidrule(lr){9-11} \cmidrule(lr){12-14} 
 && $H \uparrow$ & $Acc_r \uparrow$ & $Acc_f \downarrow$ & $H \uparrow$ & $Acc_r \uparrow$ & $Acc_f \downarrow$ & $H \uparrow$ & $Acc_r \uparrow$ & $Acc_f \downarrow$ & $H \uparrow$ & $Acc_r \uparrow$ & $Acc_f \downarrow$ \\ \midrule
Pre-train & - & - & 70.2 &  74.5& -  & 74.4 & 73.8 & - & 74.8 & 74.0 & - & 73.8 & 74.6 \\
L2$^*$& 99.73\% & 67.7 & 67.8 & 2.6 & 67.0& 65.0&4.5& 63.4 & 55.3 & 0.6 & 53.8 & 42.2 & 0.2  \\
EWC$^*$ \cite{kirkpatrick2017overcoming} &99.73\%& 69.0 & 68.8 & {1.0} & 69.2 & 67.4 &{2.6}  & 60.9 & 51.4 & {0.1} & 46.3 & 33.6 & 0.2 \\
MAS$^*$ \cite{aljundi2018memory}&99.73\% & 68.5 & 69.2 & 2.4 & 68.5 & 66.7 & 3.4 & 59.9 & 50.0 & {0.1} & 42.8 & 30.0 &{0.1}  \\
LwF \cite{feng2015learning}&99.73\% & 67.0 &67.0&3.1 &68.2&65.1&2.1 &64.0&56.1&0.3 &50.6&38.4&0.2\\
DER \cite{buzzega2020dark}&99.73\% &66.4 &67.4&4.8 &67.9&67.2&5.1&61.3&52.0&0.3&54.0&42.5&0.2\\
DER++ \cite{buzzega2020dark}&99.73\% &67.1&66.9&2.9 &68.6&67.3&3.9&63.0&54.8&0.6&64.3&57.0&0.6\\
FDR \cite{benjamin2018measuring} &99.73\%&67.2&69.5&5.0 &68.5&67.4&4.2&65.9&59.3&0.5&55.8&44.9&0.5\\
SCRUB \cite{kurmanji2023towards}&99.73\%& 67.0 & 65.5 & {1.7} & 69.2& 66.5&{1.7}& 0.0 & 0.0 & 0.0 & 18.2 & 10.4 & \textbf{0.0}  \\
SCRUB-S \cite{kurmanji2023towards}&99.73\%& 68.6 & \textbf{71.8} & {4.5} & 68.9 & \textbf{71.9} &{7.7}  & 54.8 & 63.1 & {26.4} & 19.0 & 10.9 & \textbf{0.0} \\
LIRF$^*$ \cite{Ye2022LearningWR} &50.66\% & 28.7 & 62.6 & 51.6 & 26.3 & 63.3 & 57.2 & 46.1& 54.2 & {34.7} & 46.9 & 34.9 &{2.7}  \\
Retrain &100.00\%& 13.2&7.3&\textbf{0.0}&16.2&9.1&\textbf{0.7}&13.2&7.3&\textbf{0.0}&9.5&5.1&\textbf{0.0} \\
\rowcolor{Light}
GS-LoRA &\textbf{1.28\%}& \textbf{69.3} & {70.5} & 1.9 & \textbf{71.4} & {71.1} & {2.0}  & \textbf{71.9} & \textbf{69.9} & 0.8 & \textbf{72.2} & \textbf{70.5} & 0.5 \\ \bottomrule
\end{tabular}%
\caption{\textbf{Single-step forgetting results for face recognition.} $Acc_r$ and $Acc_f$ are the accuracies of remaining and forgotten classes.
$^*$~denotes the original methods with a rehearsal buffer.
Note that ``retrain'' represents retraining the model using replay data and \textit{the training epoch is the same as other methods to ensure a fair comparison.}
Pre-train denotes the results before forgetting.
All setting is in the form of 100-Y, which means all experiments start from a pre-trained model (100 classes originally) and forget Y classes. 
}
\label{tab:single-face}
\vspace{-0pt}
\end{table*}
\begin{table*}[t!]
\centering \small
\setlength{\tabcolsep}{6pt}
\begin{tabular}{cccccccccccccc}
\toprule
\multirow{2}{*}{Methods}  &\multirow{2}{*}{\begin{tabular}[c]{@{}c@{}}Tunable\\ Ratio $\downarrow$\end{tabular}} & \multicolumn{3}{c}{80-1} & \multicolumn{3}{c}{80-5} & \multicolumn{3}{c}{80-40} & \multicolumn{3}{c}{80-70} \\ 
\cmidrule(lr){3-5} \cmidrule(lr){6-8} \cmidrule(lr){9-11} \cmidrule(lr){12-14} 
 && $H \uparrow$ & $AP_r \uparrow$ & $AP_f \downarrow$ & $H \uparrow$ & $AP_r \uparrow$ & $AP_f \downarrow$ & $H \uparrow$ & $AP_r \uparrow$ & $AP_f \downarrow$ & $H \uparrow$ & $AP_r \uparrow$ & $AP_f \downarrow$ \\ \midrule
Pre-train &-& - & 44.3 & 57.1 & - & 44.8 & 41.3 & - & 44.8 & 44.6 & - & 45.0 & 44.6 \\
L2$^*$ &99.61\%& 25.6 & 35.6 & 37.1 & 27.7 & 34.9 & 18.4 & 27.9 &32.3  &20.0  &29.9  &34.9  &18.4  \\
EWC$^*$ \cite{kirkpatrick2017overcoming} &99.61\%& 37.6 & 32.6 & 12.5 & 31.7 & 33.4 & 11.2 & 33.0 & 31.7 & 10.2 & 33.6 & 29.5 &5.7  \\
MAS$^*$ \cite{aljundi2018memory}&99.61\%&39.4  & 32.5 & 6.9 & 27.9 & 30.4 & 15.4 & 31.1 & 30.4 & 12.8 & 31.6 & 28.6 & 9.3 \\
Retrain &100.00\%& 46.6& 39.3 & \textbf{0.0} & 40.3&39.6&\textbf{0.2} & 40.5&39.2&\textbf{2.6}&37.8&39.7&8.6 \\
\rowcolor{Light}
GS-LoRA &\textbf{0.62\%}& \textbf{49.9} & \textbf{44.5} & {0.4} & \textbf{42.4} & \textbf{45.0} & {1.2} & \textbf{41.6} & \textbf{42.8} & {4.1} & \textbf{43.7} & \textbf{43.6} & \textbf{0.9} \\ \bottomrule
\end{tabular}
\caption{\textbf{Single-step forgetting results for object detection} on the COCO dataset. $AP_r$ and $AP_f$ denotes the $AP$ of remaining classes and forgotten classes.
All setting is in the form of 80-Y, which means all experiments start from a pre-trained model and forget Y classes. }
\label{tab:single-detection}
\vspace{-8pt}
\end{table*}

\noindent
\textbf{Selective Forgetting Loss.}
In each task $\mathcal{T}_t$ for $t=1,2,\cdots, T$, the model needs to forget the knowledge stored in data $D_{f_t}=(\mathcal{X}_{f_t},\mathcal{Y}_{f_t})$. 
To achieve forgetting,
the optimization goal is
$
    \arg \max\limits_\mathbf{W} \mathcal{L}\left(f_{M_{t-1}}(\mathcal{X}_{f_t}),\mathcal{Y}_{f_t}\right),
$
where $\mathbf{W}$ is the parameter;
$\mathcal{L}$ is the original loss function; 
$f_{M_{t-1}}$ is the mapping function obtained at the end of task {$t-1$}. 
An intuitive idea is to perform a negative loss, \ie,
$\mathcal{L}_{forget}=-\mathcal{L}\left(f_{M_{t-1}}(\mathcal{X}_{f_t}),\mathcal{Y}_{f_t}\right)$.
Nevertheless, simply adding a minus sign to the original loss leads to an exploding unbounded loss that is challenging to optimize. 
Therefore, we employ a ReLU function to introduce a lower bound following Du \etal \cite{du2019lifelong}, \ie,
\begin{equation}\label{eq:forget}
\mathcal{L}_{forget}=\text{ReLU}\left(\text{BND}-\mathcal{L}\left(f_{M_{t-1}}\left(\mathcal{X}_{f_t}\right),\mathcal{Y}_{f_t}\right)\right),
\end{equation}
where BND is a hyperparameter that determines the bound.

\noindent
\textbf{Knowledge Retention Loss.}
Besides forgetting selected knowledge, it is crucial for the model to maintain performance on the rest. 
Catastrophic forgetting on remaining classes \cite{kirkpatrick2017overcoming} still exists.
To mitigate this issue, we employ a small rehearsal buffer $D_{r_t}=(\mathcal{X}_{r_t},\mathcal{Y}_{r_t})$ which satisfies $|D_{r_t}|+|D_{f_t}|\ll|D|$ to alleviate this undesirable forgetting and maintain efficient training. 
The knowledge retention loss can be written as:
\begin{equation}\label{eq:remain}
    \mathcal{L}_{retain}=\mathcal{L}\left(f_{M_{t-1}}\left(\mathcal{X}_{r_t}\right),\mathcal{Y}_{r_t}\right).
\end{equation}
Combining \cref{eq:forget,eq:remain}, we get the data loss
\begin{equation}\label{eq:dataloss}
    \mathcal{L}_{data} = \mathcal{L}_{retain} + \beta \mathcal{L}_{forget},
\end{equation}
where $\beta$ is a hyperparameter.

\section{Experiments}

\subsection{Experimental Setup} \label{sec5.1}

\textbf{Datasets and Pre-trained Models.}
We evaluate the effectiveness and efficiency of GS-LoRA using published Transformer-based models in face recognition tasks and object detection tasks. 
More experiments on image classification can be found in the \textit{ Supplementary Material}.
For the face recognition task, we constructed a subdataset called CASIA-Face100 which collects 100 face IDs from the CASIA-WebFace \cite{yi2014learning} dataset.
We use a Face Transformer \cite{zhong2021face} pre-trained on the CASIA-Face100 dataset. 
For the object detection task, we use a deformable DETR \cite{zhu2020deformable} pre-trained on the COCO 2017 \cite{lin2014microsoft} dataset.

\begin{table*}[t!]
\centering\small
\setlength{\tabcolsep}{3.3pt}
\begin{tabular}{cccccccccccccccc}
\toprule
\multirow{2}{*}{Methods} & \multicolumn{3}{c}{100-20} & \multicolumn{4}{c}{80-20} & \multicolumn{4}{c}{60-20} & \multicolumn{4}{c}{40-20} \\ \cmidrule(lr){2-4} \cmidrule(lr){5-8} \cmidrule(lr){9-12} \cmidrule(lr){13-16}
 & $H \uparrow$ & $Acc_r \uparrow$ & $Acc_f \downarrow$ & $H \uparrow$ & $Acc_r \uparrow$ & $Acc_f \downarrow$ & $Acc_o \downarrow$ & $H \uparrow$ & $Acc_r \uparrow$ & $Acc_f \downarrow$ & $Acc_o \downarrow$ & $H \uparrow$ & $Acc_r \uparrow$ & $Acc_f \downarrow$ & $Acc_o \downarrow$ \\ \midrule
Pre-train & - & 74.6 & 74.6 & - & 72.9 & 70.9 & - & - & 71.9 & 69.7 & - & - & 72.7 & 71.3 & - \\
L2$^*$ & 66.7 & 61.9 & 2.3 & 63.2 & 60.9 & 5.1 & 9.4 & 63.8 & 60.3 & 2.2 & 10.1 & 62.3 & 56.7 & 2.2 &6.8 \\
EWC$^*$ \cite{kirkpatrick2017overcoming} &66.9  & 61.0 & {0.4} & 66.0 & 62.9 & {1.5} & \textbf{0.0} & 66.2 &63.9  &1.2  & \textbf{0.0} & 64.8 & 59.7 &0.5  &\textbf{0.0}  \\
MAS$^*$\cite{aljundi2018memory} & 66.6 & 60.7 & 0.7 & 65.4 & 61.8 & 1.6 & \textbf{0.0}& 66.1 & 63.5 &  {0.8}& \textbf{0.0}   & 64.2 & 58.6 &\textbf{0.3} &\textbf{0.0} \\
LwF  \cite{feng2015learning}&66.2&60.9&2.1&64.6&60.8&2.1&0.5&64.9&61.4&1.4&\textbf{0.0}&65.0&60.7&1.4&\textbf{0.0} \\
DER \cite{buzzega2020dark}&66.7&62.7&3.4&63.3&59.8&3.7&\textbf{0.0}&63.8&60.2&2.0&\textbf{0.0}&62.7&57.2&2.0&\textbf{0.0}\\
DER++ \cite{buzzega2020dark}&66.1&62.8&4.8&63.8&61.7&5.0&\textbf{0.0}&64.4&61.6&2.4&\textbf{0.0}&65.0&61.8&2.7&\textbf{0.0}\\
FDR \cite{benjamin2018measuring}&64.4&59.3&4.1&62.2&58.0&3.9&\textbf{0.0}&65.0&62.8&2.4&\textbf{0.0}&65.7&62.6&2.3&\textbf{0.0}\\
SCRUB  \cite{kurmanji2023towards}& 67.8 & 63.1 & {1.3} & 66.3 & 64.4 & {2.6} & \textbf{0.0} & 66.7 & 64.5 & {0.8} & \textbf{0.0} & 68.4 & 66.9 & 1.5 &\textbf{0.0} \\
SCRUB-S  \cite{kurmanji2023towards}&71.2  & 69.6 & {1.7} & \textbf{69.1}& {70.4} & {3.0} & {9.0} & \textbf{70.4 }&71.9 &{0.8} & {6.4} & 70.1 & 70.1 &{1.1}  &{2.3}  \\
LIRF$^*$ \cite{Ye2022LearningWR}& 28.6 & 60.1 & 55.8 & 28.3 & 58.5 & 52.2 & {43.4}& 35.8 & 56.1 &  {43.5}& {33.5}   & 36.8 & 59.8 &44.7 &22.1 \\
Retrain &18.4&10.5&\textbf{0.3}&16.0&9.1&{0.8}&\textbf{0.0}&16.9&9.6&\textbf{0.0}&\textbf{0.0}&23.4&14.0&0.5&\textbf{0.0} \\
\rowcolor{Light}
GS-LoRA & \textbf{71.6} & \textbf{72.1} & 3.5 & {68.4} & \textbf{71.1} &  4.9& \textbf{0.0} & {69.7} & \textbf{72.0} & 2.2 & \textbf{0.0} & \textbf{70.2} & \textbf{71.0} & 1.8 &\textbf{0.0} \\ \bottomrule
\end{tabular}%
\caption{\textbf{Continual forgetting results for face recognition.} $Acc_o$ is the accuracy of old tasks, \ie, the accuracy on all previously forgotten classes in task $\mathcal{T}_1,\mathcal{T}_2,\cdots,\mathcal{T}_{t-1}$.
There are 4 tasks in total and 20 classes are forgotten in each task.}
\label{tab:cl-face}
\end{table*}

\begin{figure*}[t!]
  \centering
    \begin{subfigure}{0.495\textwidth}
      \centering   
      \includegraphics[width=1\linewidth]{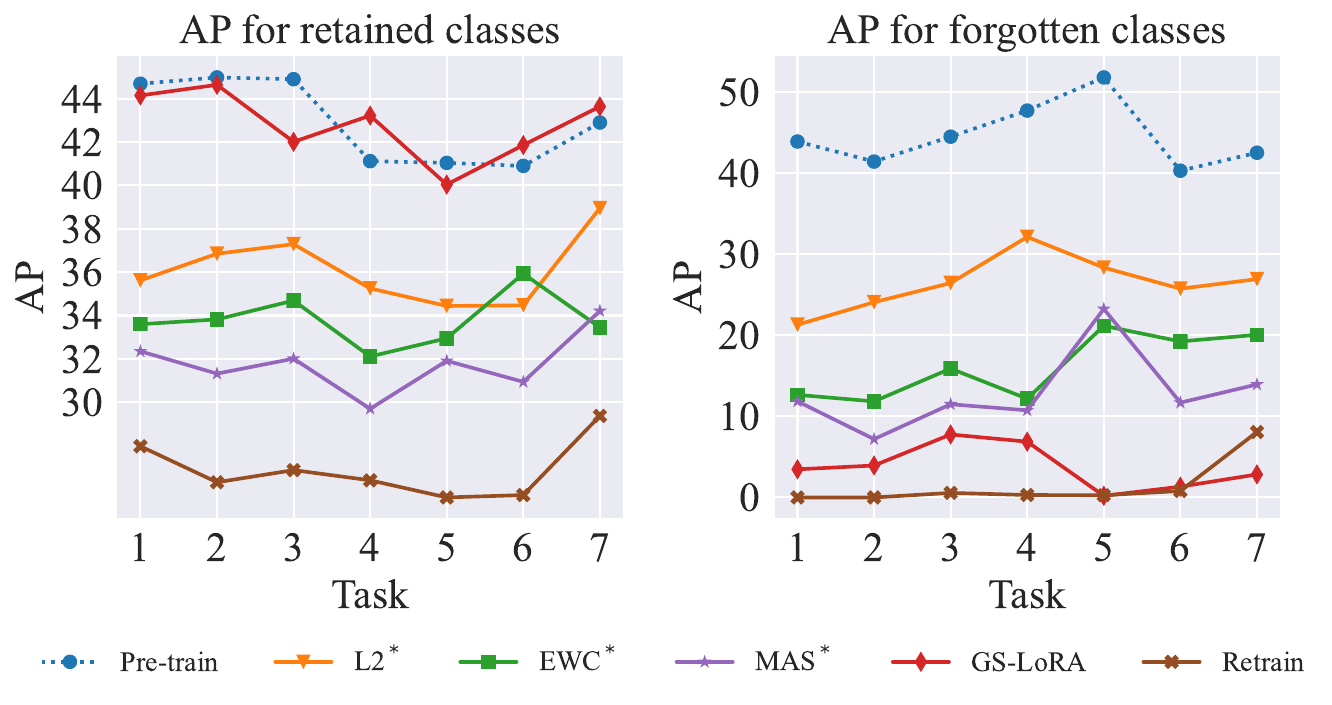}
        \caption{AP for retained  ($\uparrow$) and forgotten ($\downarrow$) classes.}
        \label{fig:sub1}
    \end{subfigure}   %
    \hfill  
    \begin{subfigure}{0.495\textwidth}
      \centering   
      \includegraphics[width=\linewidth]{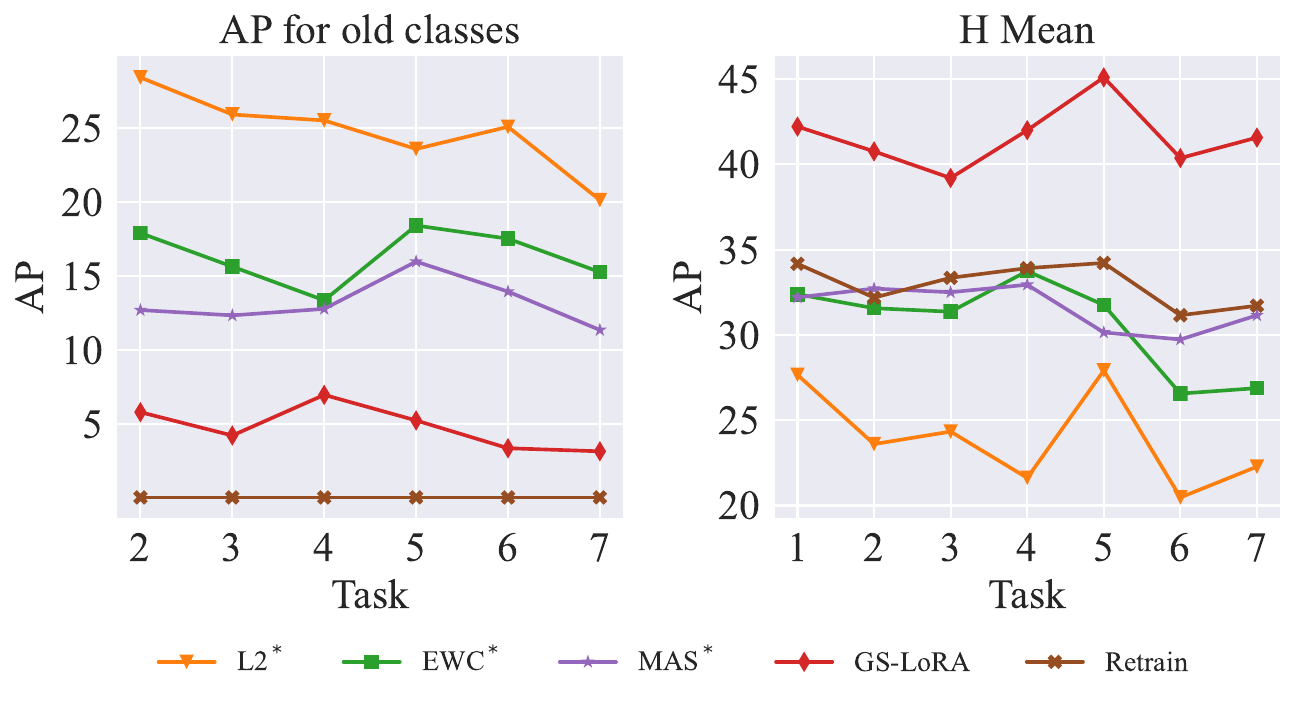}
        \caption{AP for old  ($\downarrow$) classes and H-Mean ($\uparrow$).}
        \label{fig:sub2}
    \end{subfigure}
\caption{
\textbf{Comparative results on object detection for continual forgetting.} Pre-train (\textcolor{myblue}{blue} lines) means the performance before forgetting; methods with a * indicate the original methods with rehearsal buffer.
``Retrain'' (\textcolor{mybrown}{brown} lines) refers to the process of retraining the model using replay data and \textit{the training epoch is the same as other methods for a fair comparison.}
The \textcolor{myred}{red} line is our method. There are 7 tasks in total and 10 classes are forgotten in each task.
}
\label{fig:cl-obj}
\vspace{-8pt}
\end{figure*}

\noindent
\textbf{Metrics.}
We need to evaluate the performance of the forgotten classes and the retained classes. 
We use the average accuracy (Acc) for classification tasks and the mean average precision (AP) for object detection tasks.
Ideally, the forgotten classes' performance should approach zero, and the remaining classes' performance should align with the original model's.
Similar to Shibata \etal \cite{shibata2021learning}, we define H-Mean to evaluate the overall performance after learning task $\mathcal{T}_t$, which is computed by:
\begin{equation}
    H\text{-}Mean^{(t)}=\frac{2Acc_r^{(t)} \cdot Drop^{(t)}}{Acc_r^{(t)}+Drop^{(t)}}.
\end{equation}
Here $Acc_r^{(t)}$ is calculated on the retained dataset after task  $\mathcal{T}_t$ and $Drop^{(t)}=Acc_f^{(t-1)}-Acc_f^{(t)}$ is the performance drop on forgotten classes before and after training the task.\footnote{We take the classification problem as an example to define $H\text{-}Mean$. Replace $Acc$ with $AP$ for object detection tasks.}
After learning task $\mathcal{T}_t$, we evaluate the performance on all previously forgotten classes in task $\mathcal{T}_1,\mathcal{T}_2,\cdots,\mathcal{T}_{t-1}$.

\subsection{Results and Comparisons}

We compared GS-LoRA with \textit{continual learning} methods including L2 regularization, EWC \cite{kirkpatrick2017overcoming}, MAS \cite{aljundi2018memory}, LwF \cite{li2017learning}, DER \cite{buzzega2020dark}, FDR \cite{benjamin2018measuring}, \textit{machine unlearning} methods including LIRF \cite{Ye2022LearningWR}, SCRUB \cite{kurmanji2023towards} and retraining. 
Similar to GS-LoRA, we freeze the final FFN layer to ensure backbone forgetting. 
For the retraining method, we use replay data to train a randomly initialized model and \textit{the training epoch is the same as other methods.}

\cref{tab:single-detection,tab:single-face} show the performance comparisons with the aforementioned baselines for single-task forgetting, the degraded scenario in continual forgetting.
The proposed GS-LoRA performs poorly in forgotten classes while retaining approximately the original performance in preserved classes. 
It is effective whether forgetting a small number of classes (\eg, 1 class), or a large number of classes (\eg, 90\% of all the classes).
\cref{fig:cl-obj} and \cref{tab:cl-face} show the results for continual forgetting. 
For the object detection tasks, 10 classes are forgotten per task (7 tasks in total), while for the classification task,  20 classes are forgotten per
task (4 tasks in total).
GS-LoRA works the best among the listed methods, especially on object detection tasks.
Besides, we can observe that in such a fast modification setting, severe underfitting occurs when using the retraining method.

\subsection{Ablation Study}
In this part, we conduct comprehensive ablation studies to analyze the effectiveness and efficiency of GS-LoRA.
If not specified, the default configuration for deformable DETR includes six encoders and six decoders, and the Face Transformer utilizes six Transformer blocks.
The rank we use in GS-LoRA is 8, and 10 classes are forgotten in face recognition and object detection tasks.

\begin{figure}[t]
    \centering
    \includegraphics[width=0.9\columnwidth]{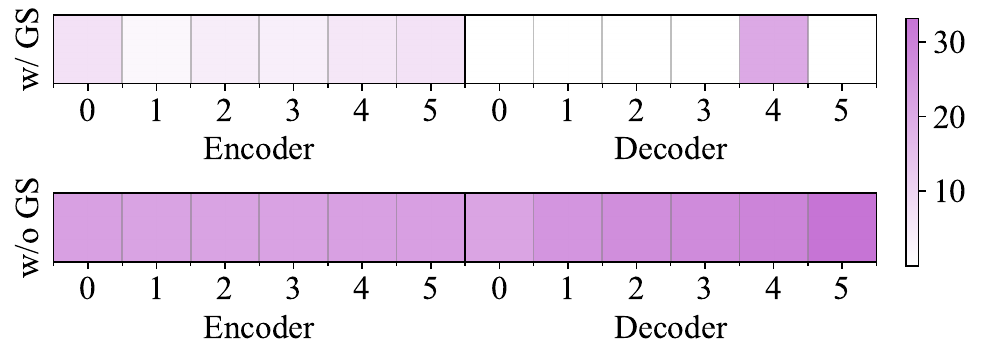}
    \caption{\textbf{Comparasion of $\ell_2$ norm of each LoRA group with or without group sparse loss.}
    Lighter colors mean smaller $\ell_2$ norms which indicate less model modification.
    The first row shows the result with group sparse loss and the second row is the result of not using it (\ie $\alpha=0$).}
    \label{fig:detr-norm}
    \vspace{-0pt}
\end{figure}

\noindent
\textbf{Group Sparsity Loss.}
We use group sparse regularization to achieve a sparser and more accurate modification.
In \cref{fig:detr-norm}, we illustrate each parameter group's $\ell_2$  norm in the deformable DETR when forgetting one class.
We can find that our GS-LoRA achieves comparable performance with directly using LoRA to fine-tune all FFNs ({forget AP: 0.40 \vs 0, remain AP: 44.49 \vs 44.51}) while requiring to modify significantly fewer parameters. 
Meanwhile, we can easily locate the knowledge more precisely with the help of the group sparsity selection strategy. 
The upper layers in the decoder contain more class-specific knowledge and need more modifications.

\begin{figure}
    \centering
    \includegraphics[width=0.9\columnwidth]{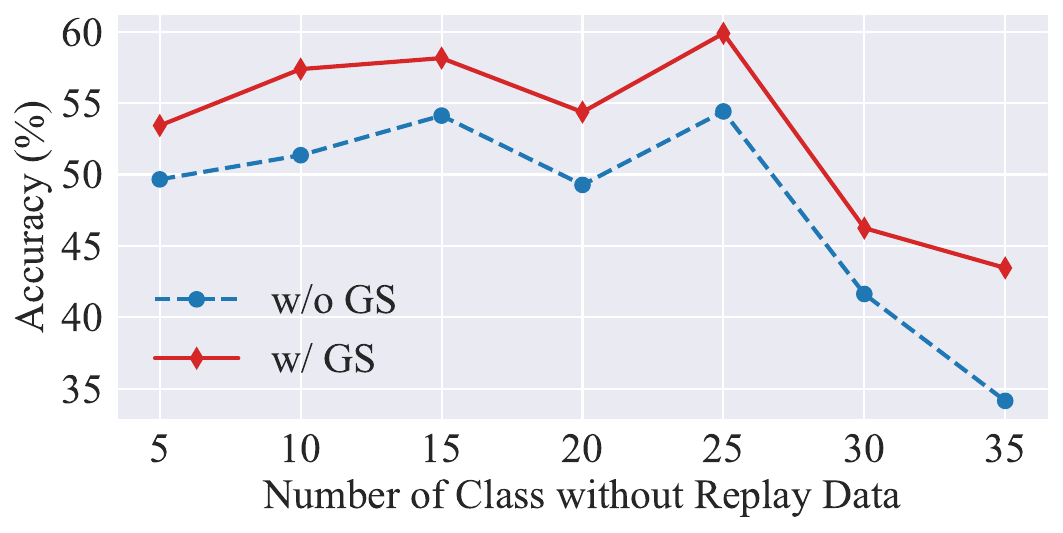}
    \caption{\textbf{Ablation study on group sparse (GS) regularization.}
In this experiment, 30 classes are forgotten. Among the remaining 70 classes, only some classes can be replayed. 
The x-axis represents the number of classes without replay data, while the y-axis denotes the accuracy of these classes.
}
    \label{fig:open}
\end{figure}

If the data of some remaining classes cannot be replayed, GS-LoRA can effectively reduce catastrophic forgetting in these classes.
We conduct the following experiments on Face Transformer.
Before forgetting, the model can identify 100 people and we want the model to forget 30 people.
\cref{fig:open} shows the results when data of certain remaining classes are not available for replay.
It is clear that GS-LoRA mitigates catastrophic forgetting on remaining classes.

\noindent
\textbf{Warm-up Sparsity.}
Illustrated by \cref{tab:alpha}, we evaluate the efficacy of our warm-up sparsity strategy on Face Transformer.
Without the warm-up sparsity, the model becomes trapped in the original local minima when $\alpha=0.01$ or larger and fails to forget.
Using a warm-up strategy, we can both achieve forgetting and easily control network sparsity.
Moreover, adopting a larger $\alpha$ can dramatically increase the network sparsity.

\noindent
\textbf{Parameter Efficiency.}
By adjusting the rank of LoRA, we can easily control the learnable parameters.
Here, we study how the rank of LoRA affects the performance when 10 classes are forgotten in the deformable DETR model.
The results in \cref{tab:rank} reveal that a larger rank tends to achieve better performance, but it will also introduce more tunable parameters.
The performance plateaus after rank goes beyond 8.
Remarkably, forgetting can be achieved using only less than 1\% of the parameters with a rank of 8.
However, most continual learning methods need to modify nearly all parameters, posing inefficiency for large models.

\noindent
\textbf{Data Efficiency.}
One benefit of our efficient forgetting paradigm is that we only utilize a small amount of data.
In practical scenarios, using too much data will dramatically increase training costs.
We compare the performance with different data ratios in \cref{tab:ratio}.
Our approach demonstrates satisfactory performance even with minimal training data, which speeds up the forgetting process. 
Although performance has a marginal improvement with increased training data, the training time rises dramatically.
\begin{table}[t!]
\centering \small
\setlength{\tabcolsep}{4.5pt}
\begin{tabular}{cccccc}
\toprule
$\alpha$ &Warm-up & $Acc_f \downarrow$ & $Acc_r \uparrow $ & $H \uparrow$ & \begin{tabular}[c]{@{}c@{}}Zero Group\\ Ratio\end{tabular}  \\ \midrule
 Pre-train& & 73.78 & 74.63 & - & - \\
 0.01& & 73.78 & 74.63 & 0.00 & 1.00 \\
\rowcolor{Light}
0.01&\usym{2713} & 1.97 & 71.06 &  71.43 & 0.17 \\
0.02& & 73.78 & 74.63 & 0.00 & 1.00 \\
\rowcolor{Light}
0.02&\usym{2713} & 0.70 & 69.99 &  71.50 & 0.50 \\
\bottomrule
\end{tabular}
\caption{\textbf{Ablation study of warm-up sparsity} on Face Transformer.
The zero group ratio is 1 means that all LoRA modules are not selected, \ie, the parameters of the pre-trained model do not change.
\textit{Without sparsity warmup, forgetting failed when $\alpha=0.01$} and $0.02$.
Zero Group Ratio is defined as the number of zero groups divided by the number of all groups.}
\label{tab:alpha}
\vspace{-8pt}
\end{table}

\begin{table}[t!]
\centering \small
\setlength{\tabcolsep}{8pt}
\begin{tabular}{cccccc}
\toprule
Rank & \begin{tabular}[c]{@{}c@{}}Tunable\\ Ratio\end{tabular}  & $AP_f \downarrow$ & $AP_r \uparrow$ & $H \uparrow$ \\ \midrule
Pre-train & -  & 43.92 & 44.73 & - \\
2 & 0.15\% & 7.60 & 44.45 & 37.98 \\
4 &  0.31\% & 6.76 & 44.33  & 40.42 \\
\rowcolor{Light}
8 & 0.62\% & 2.89 & 43.63 & 42.28 \\
16 & 1.23\% & 3.00 & 44.31 & 42.53 \\
32 & 2.47\%  & 3.26 & 44.32  & 42.40 \\ \bottomrule
\end{tabular}
\caption{{\textbf{Ablation study of the rank of LoRA modules.}}
Effective forgetting can be achieved by modifying less than 1\% parameters. }
\label{tab:rank}
\end{table}

\begin{table}[t!]
\centering\small
\setlength{\tabcolsep}{7pt}
\begin{tabular}{cccccc}
\toprule
\begin{tabular}[c]{@{}c@{}}Data\\ Ratio\end{tabular} & Speed &$Acc_f$ & $Acc_r$ & $H$ & \begin{tabular}[c]{@{}c@{}}Zero Group\\ Ratio\end{tabular} \\ \midrule
Pre-train & - & 73.78 & 74.63 & - &-\\
0.5 & 2$\times$ & 0.93 & 71.34 & 72.09 & 0.33 \\
0.2 & 5$\times$ & 1.39 & 71.29 & 71.83 & 0.50 \\
\rowcolor{Light}
0.1 & 10$\times$ & 1.97 & 71.06 & 71.43 & 0.17 \\
\bottomrule
\end{tabular}
\caption{\textbf{Data efficiency comparison.} Data ratio means the ratio of data used for forgetting to data used for pre-training.}
\label{tab:ratio}
\vspace{-8pt}
\end{table}
\section{Discussion}
\subsection{Real Forgetting or Deceptive Forgetting?}
\label{sec:6.1}

When we want to forget some specific classes, the naive solution is to mask their output FFN weights directly, which we refer to as ``head forgetting'' for simplicity.
However, this trivial solution is \textit{deceptive forgetting} and easy to be recovered.
It's like a kid who knows the answer and deliberately does not say it.
\textit{Real forgetting} should occur at backbone and is difficult to be recovered.

\begin{figure}
    \centering
    \includegraphics[width=\linewidth]{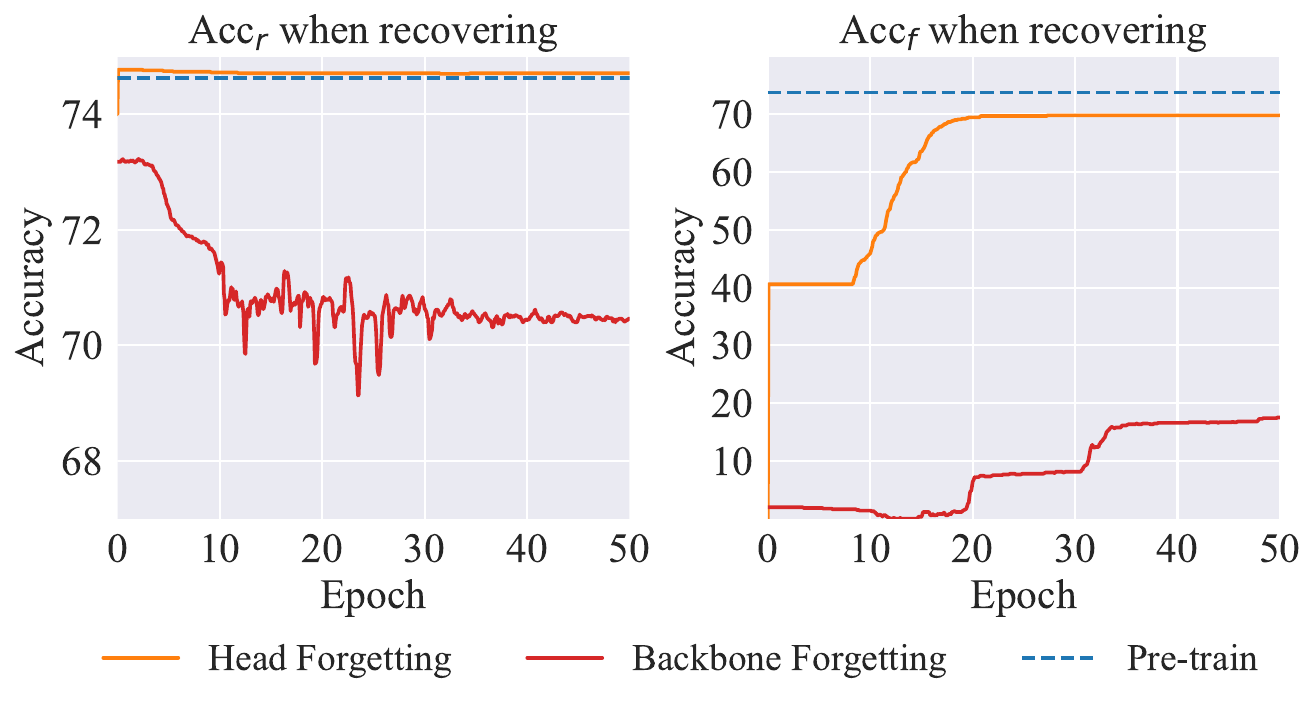}
    \caption{\textbf{Accuracy on forgotten classes and retained classes when recovering.} The \textcolor{myblue}{blue} line (Pre-train) is the result before forgetting. The \textcolor{myorange}{orange} line (Head Forgetting) is the trivial masking method. The \textcolor{myred}{red} line (Backbone Forgetting) is the GS-LoRA.}
    \label{fig:backbone}
    \vspace{0pt}
\end{figure}

We design the following experiment to demonstrate the significance of backbone forgetting.
We load a model in which forgetting has occurred, freeze its backbone, and fine-tune the output FFN layer using data containing all classes.
Then, we evaluate the model's performance on the forgotten and retained classes.
\cref{fig:backbone} shows the classification accuracy curve with epoch when recovering.
Compared to head forgetting, we can find that the model after forgetting via GS-LoRA can only be recovered to approximately 17\% on forgotten classes, significantly lower than  70\% achieved in head forgetting. 
Although it is possible to recover the accuracy of forgotten classes to a very low level in backbone forgetting, such recovery adversely impacts the accuracy of the remaining classes. 
Additionally, the recovery process for GS-LoRA needs more epochs while head forgetting can be recovered within 20 training epochs.


\subsection{Scalability}\label{sec:6.2}
We demonstrate the scalability of GS-LoRA in pre-trained models of different sizes.
We first pre-train three Face Transformer models comprising 6 blocks, 12 blocks and 18 blocks. 
It should be noted that the size of our dataset is limited and slight overfitting occurs when there are 18 blocks.
Then we use GS-LoRA to forget selective classes.
As depicted in \cref{tab:scale}, GS-LoRA exhibits remarkable scalability, demonstrating effective performance across both large and small models.
Please refer to the \textit{Supplementary Material} for visualization of group sparsity in each setting.
Combined with small tunable parameters and high data efficiency, GS-LoRA can be a useful tool for privacy erasure in large models in practice.


\begin{table}[t]
\centering\small
\setlength{\tabcolsep}{1.63pt}
\begin{tabular}{cccccccc}
\toprule
\multirow{2}{*}{\# Blocks} & \multirow{2}{*}{\# Param} & \multicolumn{2}{c}{$Acc_f \downarrow $} & \multicolumn{2}{c}{$Acc_r \uparrow $} & \multirow{2}{*}{$H \uparrow$} & \multirow{2}{*}{\begin{tabular}[c]{@{}c@{}}\# Zero\\ Groups \end{tabular}} \\  \cmidrule(lr){3-4} \cmidrule(lr){5-6}
  &  & Pre-train & Forget & Pre-train & Forget &  &  \\ \midrule
 6& 19M &73.78  & 1.97 & 74.63 & 71.06 & 71.43 & 1 \\
  12& 38M & 75.99 & 2.90 & 76.44 & 74.31 &  73.69&4  \\
 18& 57M & 73.43 & 2.55 & 73.50 & 71.20 & 71.03 &  6\\ \bottomrule
\end{tabular}
\caption{\textbf{The scalability of GS-LoRA} for three Face Transformer with different sizes. 
\# means the number of and bf stands for before forgetting, which is the result of the pre-trained model.}
\label{tab:scale}
\vspace{-8pt}
\end{table}


\section{Conclusion}



This paper presents a new and practical problem called continual forgetting and proposes an efficient and effective method to solve it.
For each continual forgetting task, we add a series of LoRA modules and only fine-tune them to achieve knowledge erasure.
Additionally, we adopt a group sparse selection strategy to select specific LoRA groups, which can make the modification more accurate and sparser.
Thorough experiments demonstrate that our method can achieve effective forgetting under various settings.

In the future, we aim to expand the applicability of our method to diverse domains, including large language models.
We believe this paper will introduce an innovative and practical direction of continual learning to the community.

\section*{Acknowledgements}
This work was supported in part by the National Key R\&D Program of China (No. 2022ZD0160102), the National Natural Science Foundation of China (No. U21B2042, No. 62320106010, No. 62072457, No. 62376267), the innoHK project, and the 2035 Innovation Program of CAS.
We thank Haochen Wang, Fei Zhu, and the CVPR reviewers for their constructive comments.

\renewcommand\thefigure{S\arabic{figure}}
\renewcommand\thetable{S\arabic{table}}  
\renewcommand\theequation{S\arabic{equation}}
\renewcommand\thesection{\Alph{section}}
\setcounter{equation}{0}
\setcounter{table}{0}
\setcounter{figure}{0}
\setcounter{section}{0}
\section*{Supplementary Material}

This document provides the supplementary materials that cannot fit into the main manuscript due to the page limit. 
Specifically, we first visualize the results on COCO dataset and the group sparsity in \cref{A}. Next, we provide more implementation details for reproducibility in \cref{B}. Finally, we provide more experiments in \cref{C}.

\section{Visualization}
\label{A}
\begin{figure*}[t!]
    \centering
    \includegraphics[width=1\linewidth]{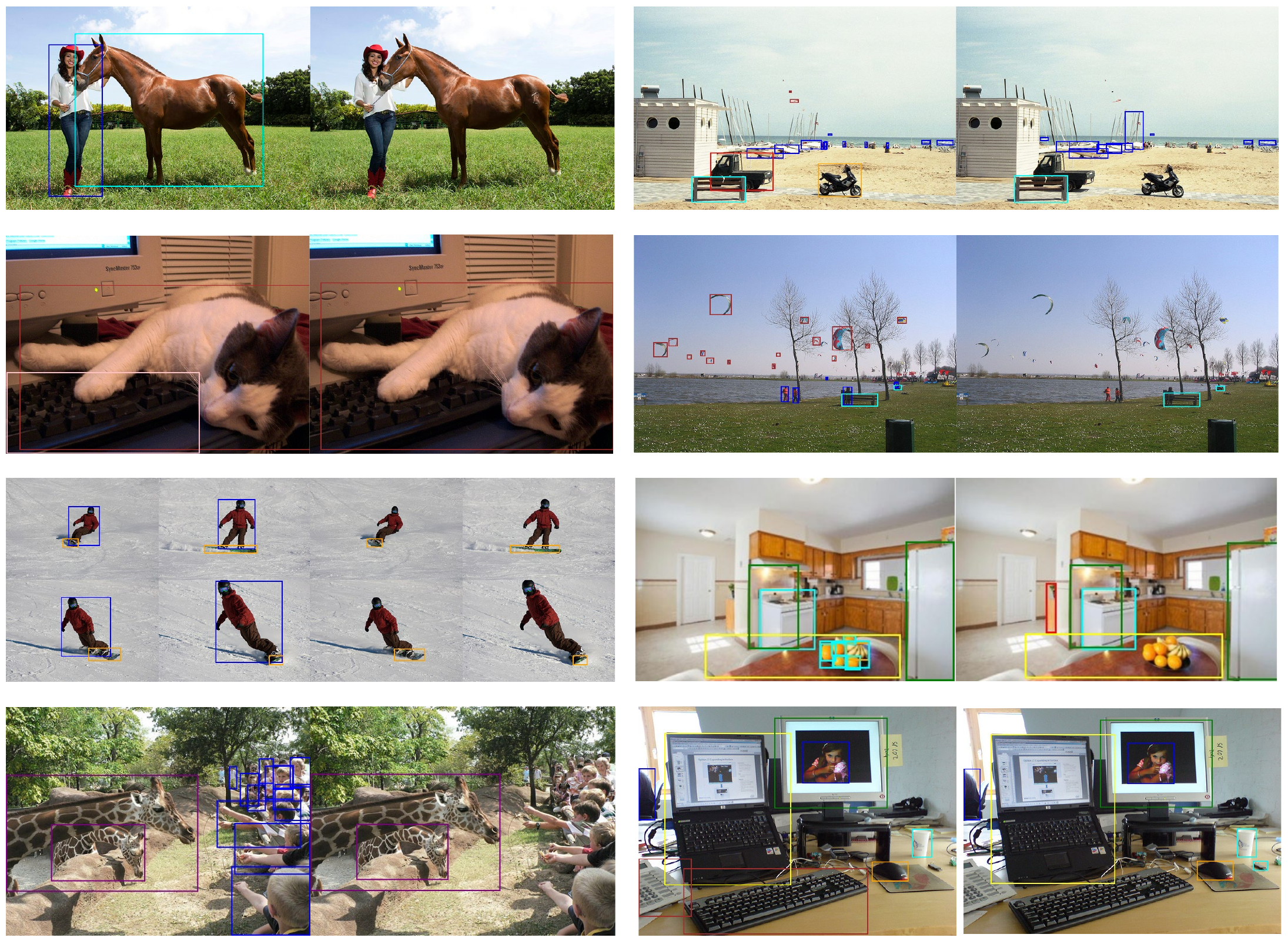}
    \caption{Visualization for \textbf{single-step forgetting}.}
    \label{fig:single-visual}
\end{figure*}

\begin{figure*}[t!]
    \centering
    \includegraphics[width=1\linewidth]{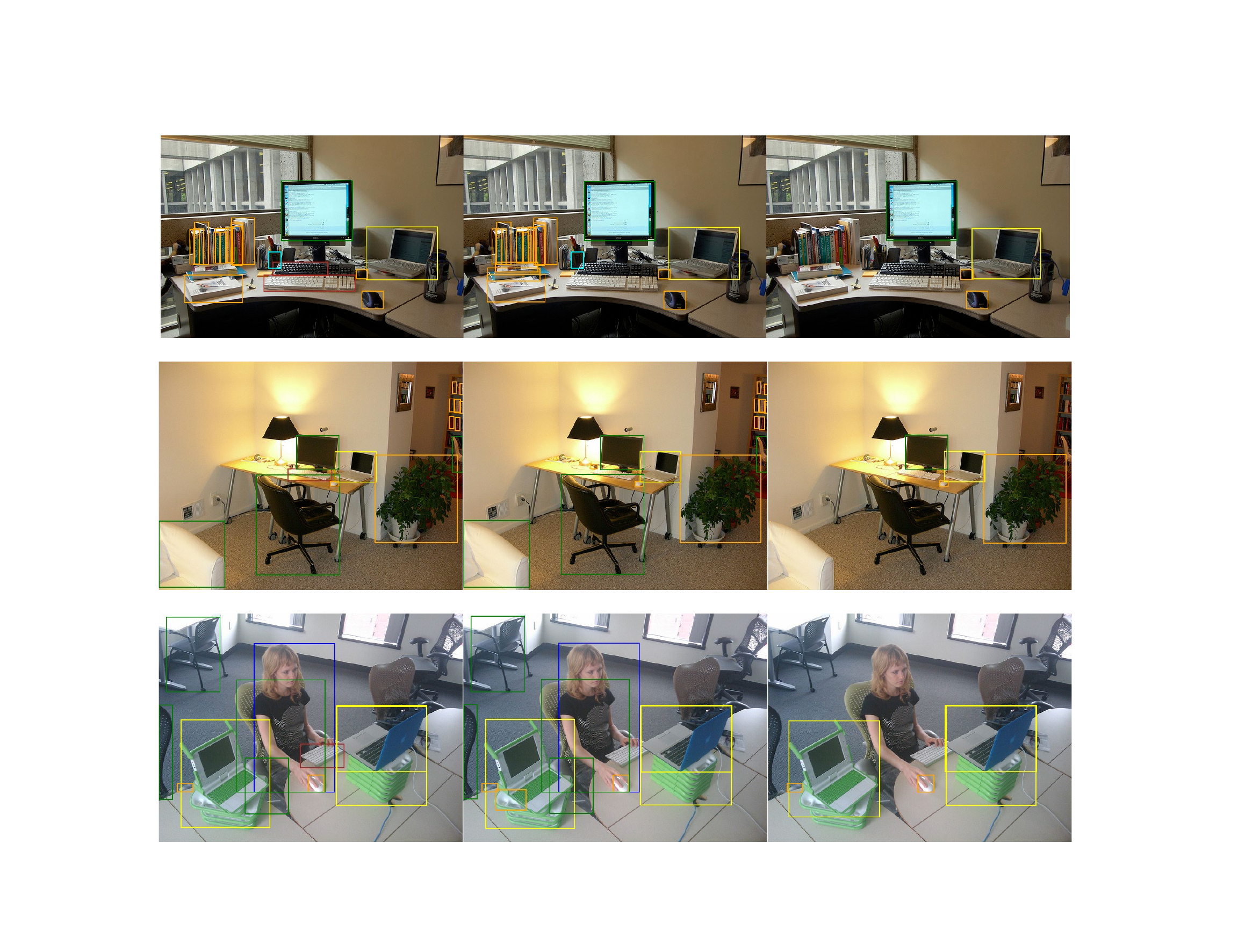}
    \caption{Visualization for \textbf{continual forgetting}. 
    The left column shows the results from the pre-trained model.
    The middle column shows the results when ``keyboard" (\textcolor{boxred}{red} bounding boxes in the left column) is erased.
    The left column shows the results when more objects (\eg, person, book, chair) are erased.}
    \label{fig:cl-visual}
\end{figure*}

\noindent
\textbf{Visualization of detection results.}
We provide visualization results on COCO validation set before and after forgetting.
\cref{fig:single-visual} is the result of single-step forgetting and \cref{fig:cl-visual} is the result of continual forgetting.
It is observed that GS-LoRA can achieve selective removal without affecting the remaining classes.

\begin{figure}[h!]
    \centering
    \includegraphics[width=\columnwidth]{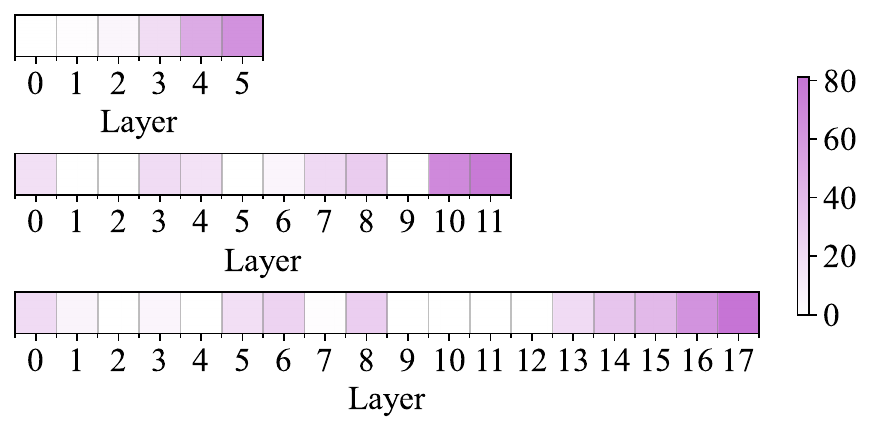}
    \caption{\textbf{$\ell_2$ norm of each LoRA group in different Face Transformers.}
    The first row shows a Face Transformer with 6 layers. The second row shows a Face Transformer with 12 layers.
    The last row shows a Face Transformer with 18 layers.
     Lighter colors mean smaller $\ell_2$ norms which indicate less  modification.}
    \label{fig:scale}
\end{figure}

\noindent
\textbf{Visualization of parameter groups.}
To show the scalability of GS-LoRA, we evaluate it on Face Transformers with 6 layers, 12 layers and 18 layers in Tab. \red{7}.
We visualize the $\ell_2$ norm of each LoRA group in these models in \cref{fig:scale}.
It is observed that GS-LoRA achieves a sparse modification on models of different sizes and shows excellent scalability.
Meanwhile, we can find that deeper layers in the Face Transformer contain more class-specific information.

\section{Implementation Details} \label{B}

\subsection{Face Recognition}
\noindent
\textbf{Network Architecture.}
Face Transformer is proposed by Zhong \etal \cite{zhong2021face} who first uses Transformer architecture to solve face recognition tasks.
A Face Transformer is a stack of Transformer blocks with a CosFace \cite{wang2018cosface} classifier.

\begin{table}[t!]
\centering\small
\setlength{\tabcolsep}{10pt}
\begin{tabular}{l|l}
Config & Value \\ \shline
optimizer & AdamW \\
base learning rate & 3e-4 \\
learning rate schedule &cosine decay \\
minimal learning rate & 1e-5\\
weight decay& 0.05 \\
momentum & $\beta_1,\beta_2=$0.9, 0.999 \\
batch size & 480 \\
warm-up epochs & 10 \\
warm-up learning rate & 1e-6\\
training epochs & 1200\\
dropout rate& 0.1\\
\end{tabular}
\caption{Pre-training settings for Face Transformer.}
\label{tab:face-pre}
\end{table}

\begin{table}[t!]
\centering\small
\setlength{\tabcolsep}{10pt}
\begin{tabular}{l|l}
Config & Value \\ \shline
optimizer & AdamW \\
base learning rate & 1e-2 \\
learning rate schedule &cosine decay \\
minimal learning rate & 1e-5\\
weight decay& 0.05 \\
momentum & $\beta_1,\beta_2=$0.9, 0.999 \\
batch size & 48 \\
warm-up epochs & 0 \\
training epochs & 100\\
dropout rate& 0.1\\
BND & 110 \\
K & 20 \\
$\beta$ & 0.15 \\
$\alpha_K$ & 0.01 \\
LoRA rank & 8 \\
data ratio &0.1\\
\end{tabular}
\caption{Forgetting settings for Face Transformer.}
\label{tab:face-forget}
\end{table}

\noindent
\textbf{Pre-training.}
We pre-train a Face Transformer on CASIA-Face100 dataset for 1200 epochs.
Implementation details can be found in \cref{tab:face-pre}.

\noindent
\textbf{Forgetting.}
For the forgetting process, we use 1337 as the random seed to generate a forgetting order.
Implementation details can be found in \cref{tab:face-forget} for all experimental settings, \textit{which shows the robustness of GS-LoRA}.
Here, BND is the bound in Eq.~(\red{10}), K and $\alpha_K$ is the hyperparameter in Eq.~(\red{9}), and $\beta$ is the hyperparameter in Eq. (\red{12}).
Note that ``data ratio" is the ratio of data used for forgetting to data used for pre-training.
To achieve fast model editing, the forgetting epoch is set to 100 and 0.1$\times$ pre-training data is used.
This is equivalent to fine-tuning the model using all pre-training data with only 10 epochs, which is \textit{less than 1\%} compared with 1200 epochs in the pre-training process.

\subsection{Object Detection}
\noindent
\textbf{Network Architecture.}
We use a Deformable DETR in object detection tasks.
Deformable DETR has 3 parts: backbone, encoder and decoder.
The backbone is an ImageNet \cite{deng2009imagenet} pre-trained ResNet-50 \cite{he2016deep}.
The encoder and decoder both have 6 Transformer blocks using multi-head deformable attention.

\begin{table}[h!]
\centering\small
\setlength{\tabcolsep}{10pt}
\begin{tabular}{l|l}
Config & Value \\ \shline
optimizer & SGD \\
momentum &0.9\\
base learning rate & 2e-4 \\
weight decay& 1e-4 \\
batch size & 16 \\
training epochs & 30\\
dropout rate& 0.1\\
BND & 15 \\
$\beta$ & 0.2 \\
$\alpha_K$ &3e-4 \\
gradient clipping & 0.1 \\
LoRA rank & 8 \\
data ratio &0.1\\
\end{tabular}
\caption{Forgetting settings for Deformable DETR.}
\label{tab:det-forget}
\end{table}

\noindent
\textbf{Pre-training.}
We use the pre-trained model released by Zhu \etal~\cite{zhu2020deformable}, where Deformable DETR is trained on COCO 2017 training set for 50 epochs and reaches 43.8 AP on COCO 2017 validation set.

\noindent
\textbf{Forgetting.}
For the forgetting process, we first generate a random list using seed 123 following Liu \etal~\cite{liu2023continual} to determine the forgetting order. Implementation details can be found in \cref{tab:det-forget} when 40 classes are forgotten.
For other experimental settings, hyperparameters are slightly different on $\beta$ and the learning rate.

\subsection{Baselines Implementation Details}
\subsubsection{Continual Learning Methods}
We implement six continual learning methods to realize continual forgetting.
Taking EWC as an example, we conduct it for forgetting as follows. 
First, we give randomly wrong labels to the forget set. 
Then, we use the remaining set to calculate the weight importance of EWC.
Finally, we regard learning on the modified forget dataset as a new task and perform EWC algorithm.
Note that we freeze the final FFN layer to ensure backbone forgetting.
Additionally, for a fair comparison, we use the remaining set as a replay buffer to enhance the performance, which is denoted as EWC$^*$.

\subsubsection{Machine Unlearning Methods}
Existing machine unlearning methods can be categorized into exact unlearning and approximate unlearning.
Exacting unlearning needs to conduct specific designs in the pre-training process, however, we cannot modify the pre-training process in a continual forgetting setting.
Initial studies on approximate unlearning are computationally heavy, \eg, \cite{guo2019certified,golatkar2020eternal,sekhari2021remember} need to calculate the Hessian matrix. 
These methods cannot be applied to large-scale problems, and we do not compare with them.
We compare our GS-LoRA with state-of-the-art LIRF \cite{Ye2022LearningWR}, SCRUB \cite{kurmanji2023towards} and SCRUB-S (a variant of SCRUB).
\begin{table*}[t!]
\centering\small
\setlength{\tabcolsep}{3.3pt}
\begin{tabular}{cccccccccccccccc}
\toprule
\multirow{2}{*}{Methods} & \multicolumn{3}{c}{100-20} & \multicolumn{4}{c}{80-20} & \multicolumn{4}{c}{60-20} & \multicolumn{4}{c}{40-20} \\ \cmidrule(lr){2-4} \cmidrule(lr){5-8} \cmidrule(lr){9-12} \cmidrule(lr){13-16}
 & $H \uparrow$ & $Acc_r \uparrow$ & $Acc_f \downarrow$ & $H \uparrow$ & $Acc_r \uparrow$ & $Acc_f \downarrow$ & $Acc_o \downarrow$ & $H \uparrow$ & $Acc_r \uparrow$ & $Acc_f \downarrow$ & $Acc_o \downarrow$ & $H \uparrow$ & $Acc_r \uparrow$ & $Acc_f \downarrow$ & $Acc_o \downarrow$ \\ \midrule
Pre-train & - & 68.0 & 60.5 & - & 64.1 & 60.9 & - & - & 69.2 & 63.3 & - & - & 76.5 & 70.7 & - \\
L2$^*$ & 47.6 & \textbf{64.9} & 22.9 & 47.6 & 68.0 & 24.3 & 30.0 & 40.0 & 70.1 & 35.3 & 24.5 & 52.7 & 74.7 & 29.9 & 23.5 \\
EWC$^*$ \cite{kirkpatrick2017overcoming} &54.1  & \textbf{64.9} & 14.2 & 55.6 & 64.9 & {12.2} & 13.9 & 53.8 & 72.7  &20.7  & 7.8 & 62.0 & 80.0 &20.1  &9.1  \\
MAS$^*$ \cite{aljundi2018memory} & 54.0 & \textbf{64.9} & 14.2 & 57.1 & \textbf{69.0} & 12.2 & 13.7 & 54.0 & 72.7 &  20.4 & 7.7   & 62.2 & 80.2 & 19.8 & 8.8 \\
Retrain &22.1&13.6&\textbf{1.0}&33.0&22.7&\textbf{0.7}&\textbf{0.0}&41.1&31.0&\textbf{2.2}&\textbf{0.0}&54.7&46.1&\textbf{3.4}&\textbf{0.0} \\
\rowcolor{Light}
GS-LoRA & \textbf{60.3} & 63.0 & 2.6 & \textbf{61.7} & 66.9 & 3.6 & 0.8 & \textbf{65.9} &\textbf{74.1} & 4.0 & 0.5 & \textbf{73.8} & \textbf{82.7} & 4.1 &\textbf{0.0} \\ \bottomrule
\end{tabular}%
\caption{\textbf{Continual forgetting results for image classification.} $Acc_o$ is the accuracy of old tasks, \ie, the accuracy on all previously forgotten classes in task $\mathcal{T}_1,\mathcal{T}_2,\cdots,\mathcal{T}_{t-1}$.
There are 4 tasks in total and 20 classes are forgotten in each task.}
\label{tab:imagenet}
\end{table*}
LIRF \cite{Ye2022LearningWR} also uses a distillation strategy to realize the deposit and withdrawal of knowledge in a model. 
For a fair comparison, we add an additional replay buffer for LIRF.

SCRUB \cite{kurmanji2023towards} uses distillation to realize efficient approximate unlearning.
The training objective is:
\begin{equation} \label{eq:SCRUB}
\begin{aligned}
 \min _{w^u} &\frac{\alpha}{N_r} \sum_{x_r \in \mathcal{D}_r} d\left(x_r ; w^u\right)\\
 &+\frac{\gamma}{N_r} \sum_{\left(x_r, y_r\right) \in \mathcal{D}_r} \ell\left(f\left(x_r ; w^u\right), y_r\right) \\
&-\frac{1}{N_f} \sum_{x_f \in \mathcal{D}_f} d\left(x_f ; w^u\right),   
\end{aligned}
\end{equation}
where 
$
d\left(x ; w^u\right)=D_{\mathrm{KL}}\left(p\left(f\left(x ; w^o\right)\right) \| p\left(f\left(x ; w^u\right)\right)\right)
$ is the KL-divergence between the student ($w_u$) and teacher ($w_o$) output distributions for example $x$, $w_o$ is the weight of a pre-trained model (teacher) and $w_u$ is the student.
$x_r$ and $x_f$ are the retained set and forgotten set which contain
$N_r$ and $N_f$ samples, respectively.
$\ell$ is the cross-entropy loss and $\alpha$ and $\gamma$ are hyperparameters.

However, Kurmanji \etal \cite{kurmanji2023towards} find that directly optimizing \cref{eq:SCRUB} is challenging and utilize a min-max optimization method following GAN \cite{goodfellow2014generative}.
To further improve the performance, we adopt a smoothing optimization method \cite{zhang2020single} as a variant of SCRUB and name it SCRUB-S.

\cref{tab:single-face,tab:cl-face} show the results in single-step forgetting and continual forgetting settings.
It is observed that LIRF cannot realize effective forgetting under our fast model erasure setting, which is data-inefficient.
SCRUB and SCRUB-S can achieve forgetting when a small number of classes need to be deleted, but GS-LoRA achieves better overall performance (H-Mean).
When we want to delete a large number of classes, \textit{only GS-LoRA can achieve complete forgetting while maintaining the performance of the rest.}
In a continual forgetting setting, SCRUB-S can achieve comparable performance with GS-LoRA, but the accuracy on previously forgotten classes ($Acc_o$) is a little bit high in SCRUB-S, which is undesirable.
In summary, the data efficiency, parameter efficiency and effectiveness of GS-LoRA make it the most applicable in real-world scenarios.

\section{More Experiments}\label{C}
In this section, we conduct more experiments to verify the effectiveness and efficiency of GS-LoRA.
In \cref{sec: unlearning}, we perform GS-LoRA in image classification tasks.
In \cref{sec:beta}, we conduct ablation studies on $\beta$ in the loss function.
In \cref{sec: incomplete}, we perform more experiments when the replay buffer is incomplete and compare GS-LoRA with continual learning baselines.

\subsection{Experiments on Image Classification} \label{sec: unlearning}
To further demonstrate the universality of our method, we use GS-LoRA to realize continual forgetting on image classification tasks.
We choose ImageNet100 \cite{deng2009imagenet} dataset and a pre-trained ViT \cite{dosovitskiy2020image} model in \ref{tab:imagenet}, where \textit{GS-LoRA still outperforms other baselines significantly}.

\subsection{Ablations on $\beta$ in Loss Function}
\label{sec:beta}
Our data loss function is \cref{eq:dataloss}, where $\beta$ controls the level of forgetting.
We conduct ablation studies in \ref{tab:beta} on face recognition tasks.
It is amazing that we find GS-LoRA demonstrates excellent performance \textit{across a wide range of $\beta$}, which shows the robustness of our method.

\begin{table}[h]
    \centering
    \small
     \setlength{\tabcolsep}{3.5pt}
     \scalebox{0.8}{
    \begin{tabular}{c|c|c|c|>{\columncolor{Light}}c|c|c|c|c|c|c|c|c}
    \toprule
       $\beta$  & 0.01 &0.05 &0.1 &  0.15 &0.2 &0.25 &0.5 &0.75 &1 &5 &10 &20 \\
       \midrule 
    $H \uparrow$ & 60.9&71.5 &71.6&\textbf{72.2}&71.8&72.0&71.5&72.0&71.8&70.1&70.5&66.3 \\
    \bottomrule
    \end{tabular}
    }
    \caption{Ablation studies on $\beta$.}
    \label{tab:beta}
\end{table}

\begin{table}[]
\centering \small
\setlength{\tabcolsep}{7pt}
\begin{tabular}{ccccc}
\toprule
\begin{tabular}[c]{@{}c@{}}Grouping\\ Strategy\end{tabular} & $Acc_f \downarrow$ & $Acc_r \uparrow$ & $H \uparrow$ & \begin{tabular}[c]{@{}c@{}}Zero Group\\ Ratio\end{tabular}  \\ \midrule
Block & 1.97 & 71.06 & 71.43 & 0.17 \\
Module & 0.93 & 70.58 & 71.70 & 0.50 \\
Matrix & 1.51 & 70.36 & 71.30 & 0.58 \\ \bottomrule
\end{tabular}
\caption{\textbf{Effect of grouping strategies.}
The zero-group ratio goes up when a more detailed grouping strategy is used.}
\label{tab:group}
\vspace{-0pt}
\end{table}

\subsection{Different Grouping Strategies}
By default, we regard two LoRA modules in a Transformer block as a group (see in \cref{fig:3}).
In this section, we explore the effect of using GS-LoRA with different grouping strategies.
In the FFN module \cite{vaswani2017attention}, there are two linear layers, each of which can add a LoRA module.
And in a LoRA module \cite{hu2021lora}, there are two low-rank matrices.

We consider three grouping strategies: \textit{``Block"}, \textit{``Module"} and \textit{``Matrix"}.
\textit{``Block"} is the default setting.
\textit{``Module"} denotes each LoRA \textbf{module} is a group, resulting in twice the number of groups compared to the Transformer blocks.
\textit{``Matrix"} means each \textbf{matrix} in LoRA modules is a group and the number of groups is four times the number of Transformer blocks.

\begin{table*}[h!]
    \centering\small
    \setlength{\tabcolsep}{7pt}
    \begin{subtable}{0.48\linewidth}
    \centering
\begin{tabular}{ccccc}
\toprule
 & $Acc_f \downarrow$ & $Acc_r \uparrow$ & $H \uparrow$ & $Acc_r^\dag \uparrow$ \\ \midrule
Pre-train & 73.67 & 75.00 & - & 73.97   \\
L2$^*$ & 0.04 & 56.81 & 64.13 & 21.06   \\
EWC$^*$ & 0.04 & 55.42 & 63.24 & 14.56   \\
MAS$^*$ & \textbf{0.00} & 55.20 & 63.11 & 16.78   \\
Retrain & \textbf{0.00 }& 14.84 & 24.70 & 5.65   \\
LoRA & 0.04 & \textbf{70.07} & \textbf{71.81} &49.66      \\
\rowcolor{Light}
GS-LoRA & 0.04 & \textbf{70.07} & \textbf{71.81} &\textbf{55.43} \\ \bottomrule
\end{tabular}
    \caption{No replay data in 5 of the 70 remaining categories.}
    \end{subtable}
    \begin{subtable}{0.48\linewidth}
    \centering
       \begin{tabular}{ccccc}
\toprule
 & $Acc_f \downarrow$ & $Acc_r \uparrow$ & $H \uparrow$ & $Acc_r^\dag \uparrow$ \\ \midrule
Pre-train & 73.67 & 75.11 & - & 75.99   \\
L2$^*$ & 0.04 & 55.61 & 63.36 & 31.12   \\
EWC$^*$ &\textbf{0.00} & 51.41 & 60.56 & 16.19   \\
MAS$^*$ & \textbf{0.00} & 51.71 & 60.77 & 17.63   \\
Retrain & \textbf{0.00} & 13.48 & 22.79 & 4.68   \\
LoRA & \textbf{0.00} & 67.57 & 70.49 &51.35      \\
\rowcolor{Light}
GS-LoRA & 0.04 & \textbf{68.49} & \textbf{70.97} &\textbf{57.37} \\ \bottomrule
\end{tabular}
    \caption{No replay data in 10 of the 70 remaining categories.}
    \end{subtable}

       \begin{subtable}{0.48\linewidth}
    \centering
           \vspace{8.5pt}
       \begin{tabular}{ccccc}
\toprule
 & $Acc_f \downarrow$ & $Acc_r \uparrow$ & $H \uparrow$ & $Acc_r^\dag \uparrow$ \\ \midrule
Pre-train & 73.67 & 74.83 & - & 76.44   \\
L2$^*$ & 0.04 & 51.65 & 60.71 & 27.32  \\
EWC$^*$ & \textbf{0.00} & 45.86 & 56.53 & 8.31   \\
MAS$^*$ & 0.04 & 47.07 & 57.43 & 12.07   \\
Retrain & \textbf{0.00} & 9.12 & 16.23 & 0.13   \\
LoRA & \textbf{0.00} & 66.64 & 69.98 &54.12    \\
\rowcolor{Light}
GS-LoRA & \textbf{0.00} & \textbf{66.85} & \textbf{70.09} &\textbf{58.14} \\ \bottomrule
\end{tabular}
    \caption{No replay data in 15 of the 70 remaining categories.}
    \end{subtable}
       \begin{subtable}{0.48\linewidth}
    \centering       \vspace{8.5pt}
        \begin{tabular}{ccccc}
\toprule
 & $Acc_f \downarrow$ & $Acc_r \uparrow$ & $H \uparrow$ & $Acc_r^\dag \uparrow$ \\ \midrule
Pre-train & 73.67 & 74.79 & - & 76.66   \\
L2$^*$ & 0.00 & 48.67 & 58.62 & 24.32   \\
EWC$^*$ & 0.00 & 41.32 & 52.95 & 6.55   \\
MAS$^*$ & 0.00 & 41.56 & 53.14 & 7.64   \\
Retrain & 0.00 & 8.14 & 14.66 & 0.10   \\
LoRA & 0.00 & 63.83 & 68.40 &49.27     \\
\rowcolor{Light}
GS-LoRA & 0.00 & \textbf{64.45} & \textbf{68.75} &\textbf{54.37} \\ \bottomrule
\end{tabular}
    \caption{No replay data in 20 of the 70 remaining categories.}
    \end{subtable}

       \begin{subtable}{0.48\linewidth}
    \centering       \vspace{8.5pt}
        \begin{tabular}{ccccc}
\toprule
 & $Acc_f \downarrow$ & $Acc_r \uparrow$ & $H \uparrow$ & $Acc_r^\dag \uparrow$ \\ \midrule
Pre-train & 73.67 & 74.81 & - & 75.88   \\
L2$^*$ & \textbf{0.00} & 44.53 & 55.50 & 22.14   \\
EWC$^*$ & \textbf{0.00} & 38.28 & 50.38 & 7.73   \\
MAS$^*$ & \textbf{0.00} & 37.81 & 49.97 & 8.24   \\
Retrain & \textbf{0.00} & 8.03 & 14.48 & 0.17   \\
LoRA & \textbf{0.00} & 64.40 & 68.72 &54.41      \\
\rowcolor{Light}
GS-LoRA & 0.04 & \textbf{66.12} & \textbf{69.68} &\textbf{59.87} \\ \bottomrule
\end{tabular}
    \caption{No replay data in 25 of the 70 remaining categories.}
    \end{subtable}
       \begin{subtable}{0.48\linewidth}
    \centering       \vspace{8.5pt}
        \begin{tabular}{ccccc}
\toprule
 & $Acc_f \downarrow$ & $Acc_r \uparrow$ & $H \uparrow$ & $Acc_r^\dag \uparrow$ \\ \midrule
Pre-train & 73.67 & 74.89& - & 74.70   \\
L2$^*$ & 0.00 & 40.86 & 52.56 & 18.77   \\
EWC$^*$ & 0.00& 33.56 & 46.11 & 5.07   \\
MAS$^*$ & 0.00 & 35.82 & 48.20 & 8.21   \\
Retrain & 0.00 & 7.13 & 13.00 & 0.50   \\
LoRA & 0.00 & 58.10 & 64.96 &41.61      \\
\rowcolor{Light}
GS-LoRA & 0.00 & \textbf{59.52} & \textbf{65.84} &\textbf{46.25} \\ \bottomrule
\end{tabular}
    \caption{No replay data in 30 of the 70 remaining categories.}
    \end{subtable}

       \begin{subtable}{0.48\linewidth}
    \centering        \vspace{8.5pt}
       \begin{tabular}{ccccc}
\toprule
 & $Acc_f \downarrow$ & $Acc_r \uparrow$ & $H \uparrow$ & $Acc_r^\dag \uparrow$ \\ \midrule
Pre-train & 73.67 & 74.82 & - & 75.02   \\
L2$^*$ & 0.00 & 37.32 & 49.55 & 18.23   \\
EWC$^*$ & 0.00 & 29.66 & 42.30 & 4.24   \\
MAS$^*$ & 0.00 & 30.04 & 42.68 & 5.66   \\
Retrain & 0.00 & 5.67 & 10.53 & 0.09   \\
LoRA & 0.00 & 52.27 & 61.15 &34.16      \\
\rowcolor{Light}
GS-LoRA & 0.00 & \textbf{56.47} & \textbf{63.93} &\textbf{43.46} \\ \bottomrule
\end{tabular}
    \caption{No replay data in 35 of the 70 remaining categories.}
    \end{subtable}
       \begin{subtable}{0.48\linewidth}
    \centering        \vspace{8.5pt}
        \begin{tabular}{ccccc}
\toprule
 & $Acc_f \downarrow$ & $Acc_r \uparrow$ & $H \uparrow$ & $Acc_r^\dag \uparrow$ \\ \midrule
Pre-train & 73.67 & 75.04 & - & 74.70   \\
L2$^*$ & 0.00 & 33.29 & 45.86 & 16.06   \\
EWC$^*$ & 0.00 & 25.08 & 37.41 & 4.15   \\
MAS$^*$ & 0.00 & 25.22 & 37.57 & 4.67   \\
Retrain & 0.00 & 6.02 & 11.13 & 0.30   \\
LoRA & 0.00 & 58.24 & 65.05 &46.48      \\
\rowcolor{Light}
GS-LoRA & 0.00 & \textbf{62.84} & \textbf{67.83} &\textbf{54.29} \\ \bottomrule
\end{tabular}
    \caption{No replay data in 40 of the 70 remaining categories.}
\end{subtable}   
    \caption{\textbf{Experiment results with incomplete replay.} Thirty classes are forgotten in all experiments. In the remaining 70 classes, only some classes can be replayed. 
    Each subtable shows the results with a different number of replay classes.
    Pre-train denotes the results before forgetting.
    L2$^*$, MAS$^*$ and EWC$^*$ denote the original methods with a rehearsal buffer.
    LoRA denotes using LoRA to fine-tune FFN modules in Transformer blocks without group sparse.
    Our method (GS-LoRA) is \colorbox{Light}{highlighted} in color.
     We specifically evaluate the accuracy of the classes without replay samples and report it as $Acc_r^\dag$.
     }
    \label{tab:more-open}
\end{table*}

We conduct our experiments on a Face Transformer and all the experiments are performed at the same sparse intensity, \ie $\alpha$ is the same. 
\cref{tab:group} shows the results of different grouping strategies. 
Notably, all grouping strategies yield exceptional performance.
It is observed that with a more detailed grouping strategy, the zero-group ratio increases, which makes sense because each group has fewer parameters and more flexibility.

\subsection{More Experiments with Incomplete Replay}\label{sec: incomplete}
In Sec. \red{5.3}, we consider the situation where we cannot obtain the replay data of some classes.
In this section, we conduct more experiments with incomplete replay data on other continual learning baselines.
We keep our settings the same as Fig. \red{5}, where 30 classes need to be forgotten.
In the remaining 70 classes, some classes cannot be replayed.
\cref{tab:more-open} shows the results when 5, 10, 15, 20, 25, 30, 35 and 40 classes cannot be replayed.
We can find that GS-LoRA mitigates catastrophic forgetting to some extent and achieves the best performance among all listed baselines.

{
    \small
    \bibliographystyle{ieeenat_fullname}
    \bibliography{main}
}


\end{document}